\def\arxivbuild{1}

\documentclass[10pt]{article}

\ifdefined\submissionbuild\def\arxivbuild{1}\fi
\newif\ifsipart
\newif\ifmainpart
\def\subpartsi{si}\def\subpartmain{main}
\ifdefined\submissionpart
  \ifx\submissionpart\subpartsi\siparttrue\fi
  \ifx\submissionpart\subpartmain\mainparttrue\fi
\fi
\ifdefined\arxivbuild
\usepackage[T1]{fontenc}
\ifdefined\submissionbuild
\usepackage[numbers,square,comma,sort&compress]{natbib}
\else
\usepackage{natbib}
\setcitestyle{authoryear,round,citesep={;},aysep={,},yysep={;}}
\fi
\usepackage{arxiv}
\expandafter\newif\csname if@accepted\endcsname
\csname @acceptedtrue\endcsname
\else
\fi

\usepackage{amsmath,amsfonts,bm}

\def\Figref#1{Figure~\ref{#1}}

\def\Secref#1{Section~\ref{#1}}

\def\eqref#1{equation~\ref{#1}}
\def\Eqref#1{Equation~\ref{#1}}

\def\1{\bm{1}}

\def\mE{{\bm{E}}}

\def\mV{{\bm{V}}}

\DeclareMathAlphabet{\mathsfit}{\encodingdefault}{\sfdefault}{m}{sl}
\SetMathAlphabet{\mathsfit}{bold}{\encodingdefault}{\sfdefault}{bx}{n}

\def\gC{{\mathcal{C}}}

\def\gE{{\mathcal{E}}}

\def\gG{{\mathcal{G}}}

\def\gL{{\mathcal{L}}}
\def\gM{{\mathcal{M}}}

\def\gV{{\mathcal{V}}}

\newcommand{\R}{\mathbb{R}}

\usepackage{amsmath}
\usepackage{amssymb}
\usepackage{booktabs}
\usepackage{array} %
\usepackage{graphicx}
\usepackage{xcolor}
\usepackage{tikz} %
\newcommand{\requested}[1]{ %
  \begin{tikzpicture}[baseline=(rq.base)]
    \node[inner xsep=4pt, inner ysep=2.2pt, outer sep=0pt,
          text height=1.45ex, text depth=0.35ex, opacity=0] (rq) {#1};
    \begin{scope}
      \clip (rq.south west) rectangle (rq.north east);
      \foreach \i in {-3,-2,...,9}{ %
        \draw[black!12, line width=4.6pt] ([xshift=\i*8pt]rq.south west) -- ++(16pt,16pt);}
    \end{scope}
    \node[inner xsep=4pt, inner ysep=2.2pt, outer sep=0pt,
          text height=1.45ex, text depth=0.35ex] at (rq.center) {#1};
  \end{tikzpicture}}
\usepackage{subcaption}
\usepackage{microtype}
\usepackage{multirow}
\usepackage{wrapfig}
\usepackage{hyperref}
\usepackage{url}
\usepackage[capitalise,noabbrev]{cleveref} %

\ifdefined\arxivbuild
\hypersetup{colorlinks=true, linkcolor=blue!50!black, citecolor=blue!50!black, urlcolor=blue!50!black,
  pdfauthor={Jonas Teufel, Clemens Binz, Pascal Friederich},
  pdftitle={Generating Graphs for a Reason: Explanation-Conditioned Adapters for Discrete Flow Matching}}
\renewcommand{\topfraction}{0.95}

\renewcommand{\textfraction}{0.05}
\renewcommand{\floatpagefraction}{0.5}
\else
\hypersetup{pdfauthor={}, pdfsubject={}, pdfkeywords={}}
\fi

\providecommand{\subauxdir}{}
\ifdefined\submissionbuild
\usepackage{xr}
\makeatletter
\let\subsavedbibcite\bibcite
\let\bibcite\@gobbletwo
\ifmainpart\fi
\ifsipart\fi
\let\bibcite\subsavedbibcite
\makeatother
\renewcommand{\headeright}{}
\crefname{appendix}{Section}{Sections}\Crefname{appendix}{Section}{Sections}
\crefname{subappendix}{Section}{Sections}\Crefname{subappendix}{Section}{Sections}
\crefname{subsubappendix}{Section}{Sections}\Crefname{subsubappendix}{Section}{Sections}
\fi

\newif\ifdraft\drafttrue
\ifdefined\arxivbuild\draftfalse\fi %
\ifdraft
  \newcommand{\todo}[1]{\textcolor{red}{\textbf{[TODO: #1]}}}
  
  \newcommand{\neednum}[1]{\textcolor{orange}{\textbf{[number: #1]}}}
  \newcommand{\note}[1]{\textcolor{blue}{\textit{[#1]}}}

\else
  \newcommand{\todo}[1]{}
  
  \newcommand{\neednum}[1]{}
  \newcommand{\note}[1]{}

\fi

\newcommand\blfootnote[1]{ %
  \begingroup
  \renewcommand\thefootnote{}\footnote{#1} %
  \addtocounter{footnote}{-1} %
  \endgroup
}

\newcommand{\phivec}{\bm{\varphi}} %
\newcommand{\phic}{\varphi_c} %
\newcommand{\nconcepts}{Q} %
\newcommand{\megan}{\textsc{Megan}}
\newcommand{\defog}{\textsc{DeFoG}}

\newcommand{\appref}[1]{Appendix~\ref{#1}}
\ifmainpart\renewcommand{\appref}[1]{Section~\ref{#1} of the Supporting Information}\fi
\ifsipart\renewcommand{\appref}[1]{Section~\ref{#1}}\fi

\title{Generating Graphs for a Reason: Explanation-Conditioned Adapters for Discrete Flow Matching}

\ifdefined\arxivbuild
\renewcommand{\shorttitle}{Generating Graphs for a Reason}
\renewcommand{\undertitle}{} %
\date{}
\newcommand{\codeurl}{https://github.com/aimat-lab/defog-megan}
\newcommand{\codedoi}{https://doi.org/10.5281/zenodo.24583646}
\newcommand{\modelsdoi}{https://doi.org/10.5281/zenodo.24583881}
\newcommand{\availabilitystatement}{ %
The code to reproduce the experiments of the main text is available at \url{\codeurl} and archived at Zenodo
(\url{\codedoi}). It includes the scripts that rebuild the datasets from the public sources. The trained models are
archived at Zenodo (\url{\modelsdoi}). The colored graphs are generated by the code. The molecular data are public,
from ZINC \citep{sterling2015zinc15}, ChEMBL \citep{zdrazil2024chembl} and the Ames benchmark \citep{hansen2009ames},
and are not redistributed.}
\author{
  Jonas Teufel \\
  Institute of Nanotechnology (INT) \\
  Institute of Anthropomatics and Robotics (IAR) \\
  Karlsruhe Institute of Technology (KIT) \\
  Karlsruhe, Germany \\
  \texttt{jonas.teufel@kit.edu}
  \And
  Clemens Binz \\
  Institute of Anthropomatics and Robotics (IAR) \\
  Karlsruhe Institute of Technology (KIT) \\
  Karlsruhe, Germany \\
  \texttt{clemens.binz@student.kit.edu}
  \And
  Pascal Friederich \\
  Institute of Nanotechnology (INT) \\
  Institute of Anthropomatics and Robotics (IAR) \\
  Karlsruhe Institute of Technology (KIT) \\
  Karlsruhe, Germany \\
  \texttt{pascal.friederich@kit.edu}
}
\else
\fi

\ifsipart
\makeatletter
\let\subpapertitle\@title
\makeatother
\title{Supporting Information for\\ \subpapertitle}
\renewcommand{\shorttitle}{Supporting Information}
\fi

\begin{document}

\maketitle

\ifsipart\else

\ifdefined\submissionbuild\else
\blfootnote{Large language models were used as general-purpose assistive
tools during the preparation of this manuscript and the accompanying
software. The authors take full responsibility for all content.}
\fi

\begin{abstract}

Generative models based on discrete flow matching can now efficiently produce
molecular graphs for a specified target property. The properties of a graph often
result from particular structural motifs within it. However, since different motifs
can lead to the same property value, conditioning on the value alone offers no
control over which of them a generated graph contains. For predictive models, such
motifs can be identified using methods from explainable artificial intelligence
(XAI), a field that aims to explain why a model makes a given prediction. In this
work, we use the explanations of a self-explaining graph neural network to inform
the generation, so that a graph can be requested not only for a target property
value but also for the structural concept through which it reaches that value. For
this purpose, we propose an adapter-based approach, in which a pretrained discrete
flow matching model is kept frozen and receives the property and the concept through
separate trainable adapters. New conditions can thereby be added to the same
pretrained generator at the cost of training one adapter each. To satisfy both
conditions at once, we additionally apply Feynman--Kac steering, which discards
unpromising candidates during generation. On a synthetic dataset with known
structure-property relationships, the adapters and steering together raise the share
of generated graphs that satisfy both conditions from 4\% for the unconditional base
model to 54\%. We believe that our approach extends the role of
explanations from interpreting predictive models to controlling generative ones,
making it possible to generate graphs that exhibit a target property \emph{for a
known structural reason}.

\end{abstract}
\ifdefined\submissionbuild
\fi

\section{Introduction}
\label{sec:intro}

\begin{figure}[t]
\centering
\includegraphics{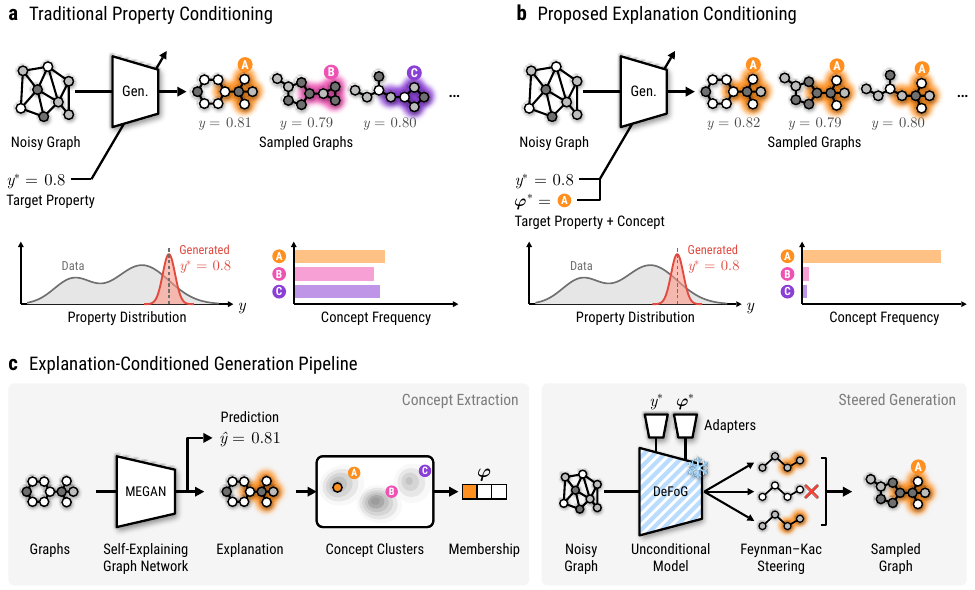}
\caption{\textbf{Overview of explanation-conditioned generation.}
\textbf{a}~Conditioned only on a target property $y^*$, a generator reaches
the target through different motifs, so that the concepts
are spread over its samples. \textbf{b}~Conditioning additionally on a target
concept membership $\phivec^*$ yields samples that reach the same target
through the requested concept, while the rest of the graph still varies.
\textbf{c}~The concepts are obtained by clustering the explanations of
\megan{}, a self-explaining graph neural network that predicts the property
and explains it in the same forward pass. Each condition enters a frozen
unconditional \defog{} model through its own adapter, and Feynman--Kac
steering discards unpromising particles during sampling. Colored highlights
mark explanations and letters identify concepts; all values are illustrative.}
\label{fig:overview}
\end{figure}

Generative machine learning models have improved substantially over the last several
years, driven in large part by diffusion and denoising models, which now produce
high-definition image and video content \citep{ho2020denoising,
rombach2022latentdiffusion, blattmann2023videoldm}. Similar advances have been made
for the generation of \textit{discrete} data structures such as \textit{graphs}. Early
approaches had to rely on continuous relaxations of the discrete graph structure
\citep{niu2020permutation, jo2022gdss}, whereas discrete flow matching
\citep{campbell2024discrete, gat2024discrete} has recently been introduced as a
simpler alternative, in which a denoising model directly predicts the transition
probabilities of a Markov chain over discrete states. Modern methods build on such
discrete formulations and achieve substantial gains in generative quality as well as
computational efficiency \citep{qin2024defog, ninniri2024freegress}. Drug design has
emerged as one of their most prominent applications, in which such models propose
molecular graphs that match a desired property profile
\citep{gomezbombarelli2018chemvae, vignac2023digress}.

Matching a property value, however, says little about the structure through which a
graph reaches it. The properties of a graph often result from particular structural
motifs within it, and since different motifs can lead to the same value, a model
conditioned on the value alone returns a mixture of them and offers no control over
which motif a generated graph contains (see Figure~\ref{fig:overview}a).

We argue that this control is a prerequisite for trust in generative design, since the
decision to act on a generated candidate ultimately rests with a human expert, for whom
the reason behind a proposal matters as much as the proposal itself. Such reasons are
the subject of explainable artificial intelligence (XAI), whose methods aim to explain
why a predictive model makes a given prediction \citep{adadi2018peeking,
yuan2023gnnexplainability}. Applied to graphs, these explanations typically highlight
the motifs on which a prediction is based, and can therefore link the structure of a
graph to its predicted property. In this work, we use such explanations to inform a
generative model for the same property, so that the motif through which a graph
reaches its property value becomes part of the request (see
Figure~\ref{fig:overview}b). In other words, instead of requesting ``generate a graph
with property X'', we request ``generate a graph with property X because it contains
motif~Y''.

Specifically, we use the notion of \textit{concept} explanations \citep{kim2018tcav,
ghorbani2019ace, koh2020concept}, which indicate the reasoning behind a model's
prediction by identifying which abstract concepts are present in the input, such as
tires and asphalt in the image of a car. On graphs, such concepts can be represented
as recurring motifs that may be strongly correlated with particular prediction
outcomes. We extract them from the self-explaining graph neural network \megan{}
\citep{teufel2023megan}, trained on the target property, and use them as additional
conditions for the generation. As the generative model, we use \defog{}
\citep{qin2024defog}, which we convert into an adapter-based variant
\citep{zhang2023controlnet, mou2024t2iadapter} with cross-attention conditioning in
its graph transformer backbone. Since the property and the concept each enter the
frozen base model through their own trainable adapter, a single unconditional model,
which can be pretrained once on a large dataset, supports individual as well as joint
conditioning, and a new condition only requires training a new adapter. To satisfy
both conditions at once, we additionally apply Feynman--Kac steering
\citep{singhal2025feynmankac}, which discards unpromising candidates during sampling.

We conduct computational experiments along a ladder of increasing realism, from a
synthetic graph dataset, in which the motifs behind every property value are known by
construction, to a molecular dataset with a deterministic property, and finally to a
real-world molecular dataset for which no independent ground truth exists. On the
synthetic graphs, the cross-attention module makes an adapter substantially more
accurate than a modulation of the transformer layers alone (decreasing the mean
absolute error from $1.08$ to $0.15$). However, the two conditions are not symmetric,
since requesting a concept also shifts the property, whereas requesting a property
value leaves the concepts as they are in the data, and we believe that this is why the
two adapters interfere when they are combined. Together, they raise the share of
generated graphs that satisfy both conditions from $0.04$ for the unconditional base
to $0.28$, and Feynman--Kac steering doubles this share to $0.54$ and almost quadruples it
for the requests that the training data rarely contains.

In summary, our contributions are as follows:
\begin{itemize}
\item We propose explanation-conditioned generation, which uses the concept
explanations of a predictive model as conditions of a generative model, so that a
graph can be requested to exhibit a target property for a known structural reason.
\item We extend \defog{} with separate cross-attention adapters for the property and
the concept on a frozen base and combine them with Feynman--Kac steering, which
together substantially improve the ability to satisfy both conditions at once.
\item We show quantitatively on synthetic graphs and qualitatively on a real-world
dataset that the approach yields graphs that satisfy both the requested property and
the requested concept.
\end{itemize}

\section{Related Work}
\label{sec:related}

\paragraph{Concept-Based and Self-Explaining Graph Neural Networks.}
Post-hoc concept-based explanation was introduced for image classifiers by
\citet{kim2018tcav} and \citet{ghorbani2019ace}, who recover a vocabulary of high-level
concepts from the internal representations of a trained model, whereas
\citet{koh2020concept} build the concept vocabulary into the model as an intermediate
bottleneck. \citet{magister2021gcexplainer} extend concept discovery to graph neural
networks by clustering node embeddings, and \megan{} \citep{teufel2023megan} instead trains an
explanation-supervised, multi-channel attention architecture whose attributional explanations
are clustered into global concepts \citep{teufel2024global}, on which we build directly
(\Secref{sec:method:megan}). In all of these methods, the concepts describe the behavior of a
predictor, and none of them uses the concept vocabulary as an input to a generative model.

\paragraph{Conditional and Guided Graph Generation.}
\citet{vignac2023digress} formulate molecular graph generation as discrete denoising diffusion
over categorical node and edge states, and \citet{qin2024defog} replace the diffusion process
with discrete flow matching \citep{campbell2024discrete, gat2024discrete}, which is the base
model that we adapt. Property conditioning follows classifier-free guidance
\citep{ho2022classifierfree, dhariwal2021guidance}, which \citet{nisonoff2025guidance} extend to
discrete state spaces, and reinforcement-learning fine-tuning optimizes a property directly
instead \citep{olivecrona2017reinvent, liu2024gdpo, rectorbrooks2025ddpp}. None of
these methods conditions on anything beyond a scalar or low-dimensional property target, so
that the mechanism by which a sample reaches the property remains unspecified.

\paragraph{Inference-Time Steering and Composing Conditions.}
\citet{wu2023tds} draw samples from a conditional distribution of a pretrained diffusion model
with sequential Monte Carlo, \citet{singhal2025feynmankac} generalize this to Feynman--Kac
steering with an arbitrary reward on intermediate steps, and \citet{li2024svdd} use a soft
value function that does not require a differentiable reward. These methods leave the model
untouched, which is why we use steering as a training-free point of comparison, but their
reward is specified from outside, whereas we derive it from the explanation model and combine
it with adapters that make the condition part of the proposal distribution. The composition of
several conditions has a long history in products of experts \citep{hinton2002poe,
du2020composable, liu2022composable, du2023reduce}, and our composition of the two adapters in
\Eqref{eq:compose} is one of this kind, whose premise of independent evidence does not hold
when one condition implies the other (\Secref{sec:discussion}).

\paragraph{Motif Vocabularies and Substructure-Constrained Generation.}
\citet{jin2018jtvae}, \citet{jin2020hiervae} and \citet{maziarz2022moler} generate molecules over
a vocabulary of structural motifs, but these vocabularies are extracted from substructure
frequency in the training data, whereas ours is extracted from what the explanations of a
trained property predictor single out as responsible for the property, which gives a
practitioner a way to ask for a structural reason behind a property value. Other methods
request a substructure directly, by extending a given scaffold \citep{lim2020scaffold}, by
decorating it with explicit attachment points \citep{arouspous2020decorator} or by linking two
given fragments \citep{imrie2020delinker}, and take the substructure as an input that the user
writes down, whereas our concept is an entry of a vocabulary that an
explanation model supplies, so that it stands for the structural reason of a property and not
for a structure that the user chose.

\paragraph{Explanation- and Concept-Guided Generation.}
Closest to our setting are methods that derive the structural request from an explanation or a
concept model. \citet{jin2020rationale} extract rationales, substructures that a property
predictor identifies as responsible for a property, expand them into full molecules with a
trained graph generator and compose the completions for several properties. Their rationale is,
like our concept, a substructure that explains a property, but it is obtained for one property
at a time and expanded by a generator that is trained for this purpose, whereas we cluster the
explanations of one self-explaining model into a global vocabulary and ask a frozen,
general-purpose generator for a concept through a lightweight adapter, jointly with the
property value. \citet{ismail2024cbgm} build a concept bottleneck into the generative model
itself, and \citet{wellawatte2022mmace} generate counterfactual molecules that explain a given
prediction by minimal edits, which explains one existing prediction post hoc, whereas our
concept adapter generates new graphs toward a structural target from the start.

\section{Method}
\label{sec:method}

\subsection{Problem Statement}
\label{sec:method:formal}

We consider undirected graphs $G = (\gV, \gE)$ with $N$ nodes, in which every node $i$
carries a categorical attribute $x_i \in \{1, \dots, A\}$ and every node pair $(i, j)$ a
categorical attribute $u_{ij} \in \{0, 1, \dots, B\}$, where $0$ indicates the absence of
an edge. Every graph can further be associated with a property function
$f : \gG \to \R$ that assigns it a scalar target value $y = f(G)$, where for a
classification task $f$ is the probability that a trained classifier assigns to the
positive class, so that a request for a class is a request for $y^* \in \{0, 1\}$. We
generally also assume that only a small number of substructures in a graph drive the value
of $f$, so that a given value $y$ can typically be reached by several different structural
configurations. Since we do not observe these substructures directly, we assume access to a
\emph{substructure membership vector} $\phivec \in [0,1]^{\nconcepts}$ that identifies which
of $\nconcepts$ recurring substructures a graph is assigned to. The generative task we
address is therefore to sample
\begin{equation}
G \sim p(G \mid y^*, \phivec^*)
\qquad \text{rather than} \qquad
G \sim p(G \mid y^*),
\label{eq:task}
\end{equation}
where $y^*$ is a requested property value, $\phivec^*$ a requested substructure membership,
and the right-hand side the property-only task of existing conditional generators. We argue
that this distinction is meaningful because $f$ is many-to-one. A target value may be
reached, for example, through a single substructure of large effect or through two
substructures of half that effect. The property-only distribution $p(G \mid y^*)$ is then a
mixture of both alternatives, weighted by how often each occurs in the training data, whereas
the additional condition $\phivec^*$ makes the choice between them part of the request.

\subsection{Multi Explanation Graph Attention Network (\megan{}) Concepts}
\label{sec:method:megan}

\megan{} \citep{teufel2023megan} is a self-explaining graph neural network $\gM$ that produces
its property prediction and an attributional explanation of it in a single forward pass,
\begin{equation}
\gM(G) = (\hat{y}, \mV, \mE),
\label{eq:megan}
\end{equation}
where $\hat{y}$ is the predicted target value and $\mV$ and $\mE$ assign every node and every
edge an importance in each of $K$ explanation channels, one for each direction of influence on
the prediction.
\citet{teufel2024global} furthermore propose an extension that extracts concept explanations
as clusters in the model's latent space of explanation embeddings, which represent only the
highlighted substructures. Density-based clustering \citep{campello2013hdbscan} per channel
yields a vocabulary of $\nconcepts$ concepts.

We translate this vocabulary into the substructure membership vector of
\Secref{sec:method:formal}. Every graph receives a vector $\phivec \in [0,1]^{\nconcepts}$,
whose entry $\phic$ is the distance of the explanation embedding of the graph to its nearest
neighbors among the members of concept $c$, sharpened by a sparsemax
\citep{martins2016sparsemax}, so that $\phic = 0$ marks core membership in the concept and
$\phic = 1$ marks the graph as unassigned to it, and we count a graph as a member if
$\phic < 0.5$. To request a single concept $c$, we use as $\phivec^*$ the mean membership
vector of its core members, so that the request has the form that the membership vector of a
real member takes.

\subsection{Discrete Flow Matching for Graph Generation (\defog{})}
\label{sec:method:defog}

\defog{} \citep{qin2024defog} casts graph generation as discrete flow matching
\citep{campbell2024discrete}, in which the attributes of \Secref{sec:method:formal} remain
categorical throughout. A noising process replaces every node and node pair by a sample from a
simple prior with probability $1 - t$ at time $t \in [0,1]$. A denoising network, a
permutation-equivariant graph transformer \citep{vignac2023digress}, receives the noisy graph
and the time $t$ and predicts the marginal distribution over the clean attribute values of
every node and node pair. Generation then runs the noising process backwards as a
continuous-time Markov chain, whose rate matrix is computed from this prediction.
The original work conditions this process by
classifier-free guidance \citep{ho2022classifierfree}, in the form that
\citet{nisonoff2025guidance} derive for discrete state spaces: the network receives the
property label $y$ for most training updates and a masked label for the remainder, so that it
learns a conditional rate matrix $R^\theta_t(\cdot \mid y)$ as well as an unconditional one
$R^\theta_t(\cdot)$, the label modulates the transformer layers through FiLM blocks
\citep{perez2018film}, and at sampling time the two are mixed as
\begin{equation}
R^{\theta,\gamma}_t(z_t, z_{t + \mathrm{d}t} \mid y)
= R^{\theta}_t(z_t, z_{t + \mathrm{d}t} \mid y)^{\gamma} \,
  R^{\theta}_t(z_t, z_{t + \mathrm{d}t})^{1 - \gamma} ,
\label{eq:cfg}
\end{equation}
where $\gamma$ is the guidance weight, with $\gamma = 1$ for ordinary conditional generation
and values above one strengthening the condition.

\subsection{Conditioning Adapters on a Frozen Base}
\label{sec:method:adapter}

We replace this conditioning mechanism by adapters over a frozen base
\citep{zhang2023controlnet, mou2024t2iadapter}. We start from a \defog{} model with
parameters $\theta$ that has been pretrained without any label, keep $\theta$ fixed
throughout, and train for every condition $c$ only the parameters $\psi$ of its adapter. Since
the base never sees a label, it can be pretrained on a large unlabeled collection of graphs,
while each adapter is trained on the smaller labeled dataset of its condition. An adapter first maps
the standardized condition $\hat{c}$ and the flow time $t$ to an embedding
\begin{equation}
\mathbf{a} = \mathrm{MLP}_{\psi}\bigl( [\, \hat{c},\, \boldsymbol{\omega}(\hat{c}),\,
\boldsymbol{\tau}(t) \,] \bigr) \in \R^{d_a} ,
\label{eq:adapter-embed}
\end{equation}
where $\boldsymbol{\tau}$ is a sinusoidal time embedding and $\boldsymbol{\omega}$ are Fourier
features that let the property adapter resolve small differences between scalar targets
(\appref{app:adapter}). The embedding then enters every layer $\ell$ of the frozen
transformer through two residual paths (Figure~\ref{fig:adapter-layer}). With $\mathbf{h}^{\ell}_i$ the hidden state of node $i$
at the output of layer $\ell$, linear heads of the adapter predict a scale
$\boldsymbol{\kappa}^{\ell}$, a shift $\boldsymbol{\delta}^{\ell}$ and a gate
$\boldsymbol{\alpha}^{\ell}$ from $\mathbf{a}$, which modulate the state as
\begin{equation}
\tilde{\mathbf{h}}^{\ell}_i = \mathbf{h}^{\ell}_i + \boldsymbol{\alpha}^{\ell} \odot \bigl(
\boldsymbol{\kappa}^{\ell} \odot \mathbf{h}^{\ell}_i + \boldsymbol{\delta}^{\ell} \bigr) ,
\label{eq:adapter-film}
\end{equation}
and the edge and graph-level states of the layer in the same way. Since this modulation applies
one diagonal affine map to every node, the adapter additionally expands $\mathbf{a}$ into $m$
condition tokens $\mathbf{t}_1, \dots, \mathbf{t}_m$, from which every node reads according to
its own state,
\begin{equation}
\mathbf{h}^{\ell+}_i = \tilde{\mathbf{h}}^{\ell}_i + W^{\ell}_O \sum_{k=1}^{m}
\operatorname*{softmax}_k \Bigl( \tfrac{1}{\sqrt{d}} \, \mathbf{q}_i^{\top} \mathbf{k}_k \Bigr)
\mathbf{v}_k ,
\qquad
\mathbf{q}_i = W^{\ell}_Q \operatorname{LN}(\tilde{\mathbf{h}}^{\ell}_i) ,\;
\mathbf{k}_k = W^{\ell}_K \mathbf{t}_k ,\;
\mathbf{v}_k = W^{\ell}_V \mathbf{t}_k ,
\label{eq:adapter-xattn}
\end{equation}
so that different nodes can receive different corrections, and $\mathbf{h}^{\ell+}_i$ replaces
$\mathbf{h}^{\ell}_i$ as the input of layer $\ell + 1$. The gates $\boldsymbol{\alpha}^{\ell}$
and the output projections $W^{\ell}_O$ are initialized to zero, so that an untrained adapter
reproduces the frozen base exactly and every adapter is a learned deviation from it. This makes
the unconditional prediction of \Eqref{eq:cfg} available without training it: it is the frozen
base itself, $p_{\theta}$. Each adapter is trained with the denoising
cross-entropy of the base, of which only $\psi$ receives gradients. We use one adapter per condition instead of a single adapter over
the concatenation of both, a property adapter conditioned on the target value $y^*$ and a
concept adapter conditioned on the target membership $\phivec^*$. The two are trained
independently and composed only at sampling time, so that each can be used alone and a new
property or a new set of explanations requires training only its own adapter.

Composing $J$ adapters at sampling time requires $J + 1$ evaluations of the denoising network
at every step, once with the frozen base alone and once with each adapter active, which for
our two conditions amounts to three evaluations from a single forward pass over a replicated
batch. Writing $p_{\theta}$ for the clean-graph marginals that the frozen base predicts and
$p_{\theta,i}$ for those predicted with adapter $i$ active, the composed prediction is
\begin{equation}
\log q = \log p_{\theta}
        + \sum_{i=1}^{J} w_i
          \left( \log p_{\theta,i} - \log p_{\theta} \right) ,
\label{eq:compose}
\end{equation}
which is normalized and from which one rate matrix is derived as in \defog{}. Every weight
$w_i$ plays the role that $\gamma$ plays in \Eqref{eq:cfg} for a single condition, so that the
composition is a product of experts in which each adapter contributes the log-ratio by which
it would move the base on its own. We sum these contributions, so that a weight of $1$ lets
every adapter move the base as far as it would alone, and the guidance strengthens with every
added condition unless the weights are lowered accordingly.

\subsection{Test-Time Steering with Feynman--Kac Guidance}
\label{sec:method:fk}

We additionally employ Feynman--Kac steering \citep{singhal2025feynmankac} as a sampling-time
intervention on top of the composed adapters. Instead of a single graph, it maintains $k$
parallel trajectories, called particles, which follow the composed model of
\Eqref{eq:compose}. At intermediate steps, every particle is scored by an energy $E$ on its
predicted clean graph, and the particles are resampled, so that promising ones are duplicated
and unpromising ones are discarded. This targets the composed distribution $p(G)$ tilted
toward low energy,
\begin{equation}
p_{\mathrm{target}}(G) \propto p(G) \exp\bigl( -\beta E(G) \bigr) ,
\label{eq:fk-target}
\end{equation}
where $\beta$ determines how strongly the energy is weighted, and the particle approximation
approaches this target as $k$ grows.

The energy expresses the joint request. For a requested concept $c$ and a requested property
value $y^*$, we use
\begin{equation}
E(G) = \phic(G) + \lambda \, \bigl( \hat{y}(G) - y^* \bigr)^2 / \sigma^2 ,
\label{eq:fk-energy}
\end{equation}
where \megan{} scores both terms, with $\phic(G)$ the membership of the graph in the requested
concept, which is zero for a core member, and $\hat{y}(G)$ its predicted property value. The
error is scaled by $\sigma$, and $\lambda$ weights the property against the concept.
\ifdefined\arxivbuild\newcommand{\expOneWidth}{0.56\textwidth}\else\newcommand{\expOneWidth}{0.5\textwidth}\fi
\newcommand{\expOneTables}{ %
\begin{wraptable}{r}{\expOneWidth}
\centering
\footnotesize
\setlength{\tabcolsep}{2.5pt}
\vspace{-1.2\baselineskip}
\caption{Property adapter on the synthetic graphs ($1024$ graphs per level in $[-3, +3]$) at guidance
weight $w = 2$, without (FiLM only) and with cross-attention (full), and the full adapter with
Feynman--Kac steering ($k = 16$ particles). MAE against the exact motif sum; best per column in bold.}
\label{tab:e1}
\begin{tabular*}{\linewidth}{l @{\extracolsep{\fill}} c @{\extracolsep{0pt}\hspace{2\tabcolsep}} c c}
\toprule
Model & MAE $\downarrow$ & Unique $\uparrow$ & Valid $\uparrow$ \\
\midrule
Base (no conditioning)        & 2.271          & 1.000 & 0.992          \\
Adapter, FiLM only            & 1.076          & 1.000 & 0.994          \\
Adapter, full                 & 0.148          & 1.000 & 0.994          \\
Adapter $+$ FK                & \textbf{0.026} & 0.812 & \textbf{0.999} \\
\bottomrule
\end{tabular*}

\vspace{1.0\baselineskip}
\setlength{\tabcolsep}{3pt}
\caption{What each adapter changes on the synthetic graphs. Match is the share of graphs that meet a
requested condition, a property within $\tau = 1$ of a level or a membership $\phic < 0.5$ in a
concept, averaged over the seven levels and the $19$ concepts. Shift is how far a condition the adapter
was not asked for deviates from what the data imply, in percent of its range. The hatched cells are the
requested condition.}
\label{tab:e1-effects}
\begin{tabular*}{\linewidth}{@{\extracolsep{\fill}} l cc cc}
\toprule
 & \multicolumn{2}{c}{Match $\uparrow$} & \multicolumn{2}{c}{Shift (\%) $\downarrow$} \\
\cmidrule(lr){2-3}\cmidrule(lr){4-5}
 & property & concept & property & concept \\
\midrule
Base (no conditioning) & 0.39 & 0.06 & ---  & ---  \\
Property adapter       & \requested{0.87} & 0.06 & ---  & 1.2  \\
Concept adapter        & 0.39 & \requested{0.72} & 20.3 & ---  \\
\bottomrule
\end{tabular*}
\vspace{-0.8\baselineskip}
\end{wraptable}
}
\section{Computational Experiments}
\label{sec:experiments}
\setcounter{topnumber}{1}
\ifdefined\arxivbuild\expOneTables\fi

We evaluate the method along a ladder of increasing realism: synthetic colored graphs with a
known structural mechanism (Experiments~1 and~2), a molecular dataset with a deterministic
property (Experiment~3) and a real-world molecular dataset without independent ground truth
(Experiment~4), each time also against Feynman--Kac steering \citep{singhal2025feynmankac} as a
training-free alternative.

\paragraph{Synthetic Baseline and Individual Adapter Guidance.}
In this experiment, we first investigate the two proposed adapters, for the property and for the
concept membership, individually on a dataset of synthetic graphs whose underlying mechanism is
known. For this purpose, we use a version of the RbMotifs dataset \citep{teufel2023megan}, a
synthetic graph regression dataset in which different red and blue motifs are deliberately seeded
into random colored graphs to induce a positive and a negative contribution to the ground truth
value assigned to the graph respectively. By construction, this property is many-to-one, since
multiple combinations of positively and negatively influencing motifs result in the same graph
value. We train a \defog{} base model on these graphs without the labels to obtain an
unconditional generative model, and we train a \megan{} model to predict the graph values and
their attributional explanations. We furthermore derive the concept membership vectors from this
trained model by the clustering in its latent space described in \Secref{sec:method:megan}. Both
adapters for the generative model are then trained with the same graph values and concept
memberships as the conditions (\appref{app:data:rbmotifs}).

\ifdefined\arxivbuild\else\expOneTables\fi

Used individually for their respective guidance purposes, the two adapters first of all confirm
the effectiveness of the proposed extensions of the original \defog{} framework. Comparing the
FiLM-based property conditioning of the original work with our cross-attention adapter, we
observe a sevenfold reduction of the error (Table~\ref{tab:e1}). Combining this adapter with
additional inference-time Feynman--Kac guidance reduces the property error by another factor of
five, while still maintaining a uniqueness of $81\%$. We observe a similar improvement for the
concept adapter, which raises the match rate from $6\%$ to $72\%$
(Table~\ref{tab:e1-effects}), and we report the detailed results per level and per concept in
\appref{app:perconcept}.

Running the individual guidance scenarios reveals another important aspect of the task. While
controlling the property largely leaves the distribution of the concepts unaffected (a shift of
$1.2\%$, Table~\ref{tab:e1-effects}), controlling for a concept substantially shifts the property
distribution of the generated graphs (a shift of $20\%$). Ultimately, this is the expected
behavior, since an explanatory concept is itself defined by a concrete effect on the property, so
that the members of a concept are expected to have a biased property distribution. In practice,
however, this means that the two tasks are not independent, and perfectly controlling the
property to an arbitrary range may be considerably harder depending on how strongly the property
distribution of a given concept is skewed.

\paragraph{Joint Conditioning on Synthetic Graphs.}
Following the individual guidance, this experiment investigates the feasibility and the
efficiency of joint guidance for the synthetic RbMotifs dataset. For this purpose, we generate
graphs with both adapters active and analyze the individual as well as the joint match rate. For
the property we choose five value levels $\{-3, -1, 0, +1, +3\}$, which cover the whole property
range, and for the explanations we choose a subset of six representative concepts, which yields
$30$ target cells with $1024$ graphs each. Here we define the joint match rate such that a
generated graph has to be within $\tau = 1$ of the desired property and has to be recognized as a
member of the requested concept by the \megan{} model. We compare the composed adapters at the
weights $(w_c, w_p) = (1, 1)$, against the
unconditional base, each adapter alone, and Feynman--Kac steering with $k = 16$ particles, which
we apply both to the composed adapters and to the base alone. A sweep of weights and steering
strengths shows the same ordering of the arms over most of the range (Figure~\ref{fig:e2} in
\appref{app:fk_adapter}).

\begin{table}[t]
\centering
\small
\setlength{\tabcolsep}{3.5pt}
\caption{Joint conditioning on the synthetic color graphs for $30$ target cells
($1024$ graphs per cell). A joint match requires the exact property within $\tau = 1$ of
the target and core membership in the requested concept, the native and off-native
columns give the joint match per cell class, depending on whether at least $15\%$ of the
concept's core members carry the requested property, and cost is the median number of samples
per distinct joint match over the $27$ supported cells. The best result per column is
highlighted in bold.}
\label{tab:e2}
\begin{tabular*}{\textwidth}{l @{\extracolsep{\fill}} c @{\extracolsep{0pt}\hspace{2\tabcolsep}} cc cc cc r}
\toprule
 & \multicolumn{3}{c}{Match $\uparrow$} & \multicolumn{2}{c}{Joint by cell $\uparrow$} & & & \\
\cmidrule(lr){2-4}\cmidrule(lr){5-6}
Model & Prop. & Concept & Joint & Native & Off-nat. & Unique $\uparrow$ & Valid $\uparrow$
      & Cost $\downarrow$ \\
\midrule
Base (no conditioning)       & 0.392          & 0.119          & 0.044          & 0.086          & 0.027          & \textbf{1.000} & 0.995          & 25.6 \\
$+$ property adapter         & \textbf{0.956} & 0.119          & 0.114          & 0.298          & 0.026          & \textbf{1.000} & 0.993          & 26.3 \\
$+$ concept adapter          & 0.321          & \textbf{0.749} & 0.222          & 0.540          & 0.074          & \textbf{1.000} & 0.991          & 21.8 \\
$+$ both adapters            & 0.541          & 0.500          & 0.276          & 0.646          & 0.107          & \textbf{1.000} & 0.993          & 7.2 \\
$+$ FK ($k{=}16$)            & 0.873          & 0.481          & 0.405          & 0.690          & 0.309          & 0.760          & \textbf{0.999} & 2.7 \\
$+$ both adapters $+$ FK     & 0.732          & 0.741          & \textbf{0.544} & \textbf{0.932} & \textbf{0.412} & 0.817          & \textbf{0.999} & \textbf{2.2} \\
\bottomrule
\end{tabular*}
\end{table}

The results of the experiment indicate that running both adapters independently is the most
effective configuration to maximize their respective individual match rates ($96\%$ for the
property steering and $75\%$ for the concepts, Table~\ref{tab:e2}). Applying both adapters at the
same time results in the highest joint match rate of the trained arms, $28\%$, but lowers the
individual match rates to only about $50\%$ each. Interestingly, purely inference-time
Feynman--Kac steering alone achieves an even higher joint match rate of $41\%$. The best
configuration, however, uses both adapters and Feynman--Kac steering at the same time and achieves a joint
match rate of $54\%$.

To further illustrate the efficacy of the joint guidance, we calculate the conditional sampling
cost as the number of samples that have to be drawn from a given configuration to obtain one
unique graph with the desired combination of property and concept. This takes the joint match
rate into account, but also the validity and the uniqueness of the samples. For the unconditional
base model it evaluates to about $26$ samples, while the best combined configuration lowers it to
a median of $2.2$ samples.

\begin{figure}[t]
\centering
\includegraphics{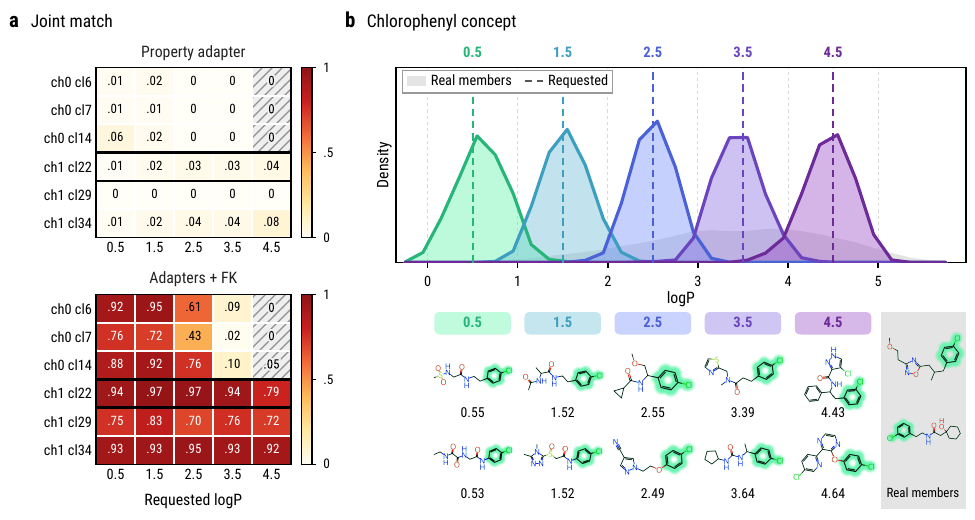}
\caption{\textbf{Steering a concept and the $\log P$ of the molecule at the same
time.} \textbf{a}~Joint match per concept and requested $\log P$ level for the property adapter
alone (top) and for the composed adapters with steering (bottom); each cell is the fraction of
$1024$ generated graphs that satisfy both conditions, counting repeated molecules, hatched cells have
no training molecule with that concept at that level, and the frame marks the concept of panel~b
(Table~\ref{tab:e4} gives the averages). \textbf{b}~Density of the $\log P$ of the steered molecules
that contain the chlorophenyl concept (ch1 cl22), one curve per requested level (dashed lines), over
its real training members (grey), with two example molecules per level chosen by eye and two real
members on the grey block. The highlighted group is a substructure match, not the explanation of the
model.}
\label{fig:e4}
\end{figure}

\begin{table}[t]
\centering
\small
\setlength{\tabcolsep}{3.5pt}
\caption{Joint conditioning on molecular graphs for $30$ target cells ($1024$ graphs per cell): the
numbers behind Figure~\ref{fig:e4}, with the columns of Table~\ref{tab:e2}. The best result per column is
highlighted in bold.}
\label{tab:e4}
\begin{tabular*}{\textwidth}{l @{\extracolsep{\fill}} c @{\extracolsep{0pt}\hspace{2\tabcolsep}} cc cc cc r}
\toprule
 & \multicolumn{3}{c}{Match $\uparrow$} & \multicolumn{2}{c}{Joint by cell $\uparrow$} & & & \\
\cmidrule(lr){2-4}\cmidrule(lr){5-6}
Model & Prop. & Concept & Joint & Native & Off-nat. & Unique $\uparrow$ & Valid $\uparrow$
      & Cost $\downarrow$ \\
\midrule
Base (no conditioning)       & 0.192          & 0.014          & 0.003          & 0.004          & 0.001          & \textbf{1.000} & 0.988          & 1024.0 \\
$+$ property adapter         & 0.844          & 0.017          & 0.015          & 0.022          & 0.007          & \textbf{1.000} & 0.946          & 113.8 \\
$+$ concept adapter          & 0.183          & 0.357          & 0.067          & 0.102          & 0.028          & \textbf{1.000} & 0.978          & 19.0 \\
$+$ both adapters            & 0.698          & 0.264          & 0.194          & 0.223          & 0.203          & \textbf{1.000} & 0.948          & 4.4 \\
$+$ FK ($k{=}16$)            & 0.871          & 0.091          & 0.064          & 0.096          & 0.031          & 0.338          & \textbf{0.999} & 128.0 \\
$+$ both adapters $+$ FK     & \textbf{0.932} & \textbf{0.719} & \textbf{0.676} & \textbf{0.813} & \textbf{0.639} & 0.365          & 0.992          & \textbf{4.2} \\
\bottomrule
\end{tabular*}
\end{table}

\paragraph{Joint Conditioning on Molecular Graphs (Crippen $\log P$).}
We repeat the joint conditioning with a more realistic experiment on molecular graphs, using the
Wildman--Crippen approximation of the octanol-water partition coefficient
\citep{wildman1999clogp} as the target property, which can be calculated deterministically with
the popular cheminformatics software RDKit \citep{rdkit}. For this purpose, the frozen generative
base is pretrained on the ZINC-250k dataset \citep{sterling2015zinc15}. We train both adapters
and the \megan{} model on a subset of about $65$k labeled ZINC molecules and the resulting $42$
concept clusters (\appref{app:data:zinc}).
We evaluate this task using five levels of $\log P \in \{0.5, \dots, 4.5\}$ and six
representative concepts ($1024$ graphs per cell), where a joint match requires a valid molecule
whose Crippen $\log P$ lies within $0.5$ of the target.

The results reflect the increased difficulty of this task when compared to the synthetic graphs.
The unconditional base model has a joint match rate of essentially $0\%$, which makes it
extremely unlikely to randomly stumble upon a molecule with an arbitrary combination of concept
and property. The best configuration, again, uses a combination of both adapters and Feynman--Kac steering
and results in a joint match rate of $68\%$, although at a significantly higher cost to the
uniqueness of the resulting samples (only about $36\%$ unique samples, Table~\ref{tab:e4}).
Overall, the best combination achieves a conditional sampling cost of about four samples until a
desired combination is matched exactly. We attribute this loss of diversity to the much smaller
space of valid molecules when compared to the randomized synthetic graphs, since chemical valency
rules only allow very specific combinations of atoms and bonds.

As already implied by the individual steering experiments, we also find that the steerability to
certain regions of the property space depends on the given concept. While some concepts, such as
the chlorophenyl motif, can be effectively steered across the whole range of property values,
other concepts can only be reached within a more limited range (Figure~\ref{fig:e4}a; the
concept ch0 cl7, for example, is matched in about three quarters of the samples at the two
lowest levels and almost never at the two highest). On the one hand, certain combinations of
concept and property value practically do not exist in the dataset at all. On the other hand,
certain combinations might not be physically possible, and we believe that this is the stronger
limit of the two. If a concept represents a strongly polar functional group, for instance, a very
high partition coefficient may be out of reach no matter what the rest of the molecule looks
like, since the group itself pulls the value down.

\begin{figure}[t]
\centering
\includegraphics{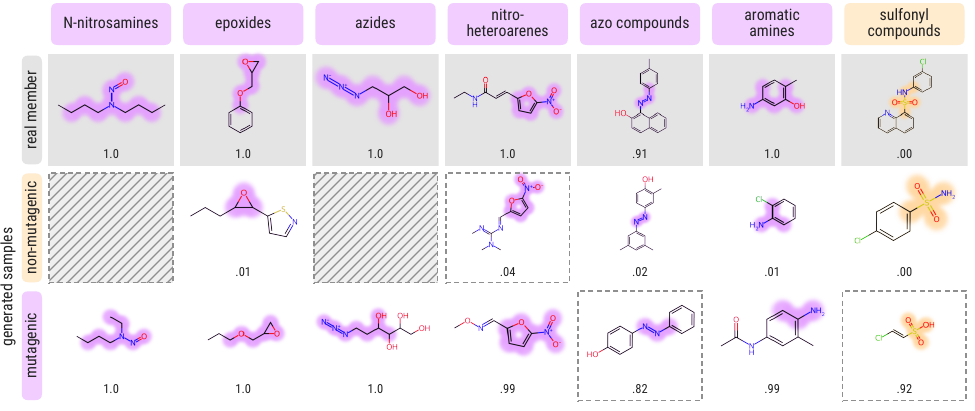}
\caption{\textbf{Example molecules generated for a requested concept and class of
mutagenicity.} Each column is a concept of \megan{}, named by the substructure that
dominates its core members, with a real member (grey) and one molecule generated
with both adapters and steering for the non-mutagenic (orange) and the mutagenic
class (purple).
Highlighted atoms are the explanation of \megan{} in the channel of the concept
(color of the box behind its name), and the number is the probability of
mutagenicity that \megan{} predicts, not a measurement. Cells with fewer than ten
core members at the requested class lie beyond the data and are framed by dashes,
and hatched when no molecule was found.}
\label{fig:e5}
\end{figure}

\paragraph{Qualitative Experiments on Real-World Mutagenicity Data.}
The previous experiment used the Crippen $\log P$, which has the downside of being a simplified
surrogate for an experimentally measured partition coefficient, but the upside of offering a
cheap ground truth oracle for every generated molecule. For the majority of real-world
properties, and for experimental ones in particular, this is not the case. In this experiment, we
therefore investigate the property of mutagenicity, the predisposition of a substance to mutate
or otherwise harm the DNA of living organisms. For this purpose, we train a second frozen base
model on about one million molecules of the ZINC \citep{sterling2015zinc15} and ChEMBL
\citep{zdrazil2024chembl} databases (\appref{app:data:corpus}), which adds the charged atom classes that nitro, azide and
N-oxide groups require, and we train the property adapter as well as the \megan{} model on the
Ames mutagenicity dataset of \citet{hansen2009ames} (\appref{app:data:ames}). This dataset presents a classification task,
in which every molecule is labeled as either mutagenic or non-mutagenic based on experimentally
conducted Ames assays with the bacterium \textit{Salmonella typhimurium}.

In the absence of a ground truth oracle, we can only interpret the results qualitatively, and
their classification is derived from the trained \megan{} model. For the mutagenicity task, each
concept explanation is associated with one of the two classes, so that requesting a concept
already biases the class distribution of the generated molecules strongly. This is to be
expected, since previous work has established that the concept explanations of the mutagenic
channel correspond to structural alerts that are already known for mutagenicity
\citep{teufel2024global, kazius2005toxicophores}. Regardless, the generative model produces
molecules that carry a concept of the mutagenic channel and that the model still classifies as
non-mutagenic.

Structural alerts are often used as a proxy for mutagenicity, but the behavior is usually more
complex than the presence of a group. One example is the aromatic amines (the sixth column of
Figure~\ref{fig:e5}), which are activated to a nitrenium ion by $N$-hydroxylation, where the
stability of that ion depends on the whole ring system \citep{benigni2011mechanisms,
muto2024nitrenium}. In the molecule generated as mutagenic, the free amine sits across the ring
from an acetamido group that can stabilize the ion, whereas the molecule generated as
non-mutagenic places the amine next to a chlorine atom, a pattern that is generally associated
with lower mutagenic activity (\appref{app:nitrenium}). This example indicates that a generated molecule can carry a
chemically sensible reason for the class that was requested, even on a task for which no oracle
exists. However, such a reading remains a conjecture. Without a ground truth, a molecule that
looks sensible may equally be a misclassification of a predictive model whose accuracy is
limited ($0.76$ on held-out compounds).

\setcounter{topnumber}{2} %

\section{Discussion and Limitations}
\label{sec:discussion}
\label{sec:limitations}

One important finding of the preceding experiments is that an adapter-based approach can achieve
the individual as well as the jointly guided generation of graphs that exhibit a target property
and a target explanatory motif. Compared with the classifier-free guidance of the original
model, such an approach offers a considerably higher degree of flexibility. A single
unconditional model can be trained on a large library of unlabeled graphs, while each adapter is
trained on the much smaller, specialized dataset of its own condition, and separate adapters
make it possible to mix and match different property and concept conditions. We expect this
modularity to come at the cost of conditioning quality in general, but we are able to counteract
much of this loss with the cross-attention conditioning and the additional inference-time
Feynman--Kac steering (\appref{app:fk_adapter:native}).

Furthermore, the experiments confirm that the concept guidance can itself induce a strong shift
in the property distribution, which follows from the very nature of the concept explanations and
depends on the requested concept. For certain properties, it may moreover be physically
near-impossible to reach parts of the property range for a given target concept at all, no
matter how the rest of the graph is chosen. Whether a combination is out of reach for physical
reasons or only for want of data, this presents a limitation of the proposed approach, and
especially so because it is not evident in advance which combinations of concept and property
range are the most difficult for the model.

One additional general limitation of the proposed method concerns the nature of the concept
explanations themselves. They are derived from a model, and they are therefore approximations
that rest on assumptions about the underlying property. Several training runs on the same
dataset converge on similar, yet slightly different concept explanations. In addition, the
datasets on which such models are fitted are often small, biased in the region of chemical space
that they cover, and subject to experimental noise. In summary, the concept conditioning in
particular requires a multi-step procedure in which process and approximation noise enter at
several points, which may lead to an unfavorable accumulation of errors overall.

\section{Conclusion}
\label{sec:conclusion}

In this work, we present a method for conditioning graph generation jointly on a target
property value and a target concept explanation, so that a generated graph can be asked to
exhibit a property for a specific structural reason. To that end, we extend the \defog{}
architecture for discrete flow matching with an adapter-based approach to conditioning, and we
use separate property and concept adapters to jointly steer the generation of a frozen
unconditional base model. We furthermore improve the conditioning results with a cross-attention
conditioning module and with inference-time Feynman--Kac steering.

On a synthetic dataset of colored graphs with known mechanisms and on a molecular dataset with a
simplified property, our experiments show that each adapter on its own steers the generation
effectively. We find that sampling from the joint distribution is most effective with a
combination of both adapters and inference-time Feynman--Kac steering, at a cost to the
diversity of the samples. We furthermore demonstrate the approach qualitatively on a real-world
mutagenicity dataset, for which no ground truth oracle exists.

In general, we believe that the proposed method presents a first step toward a greater degree of
flexibility and usability for graph generation, and for molecule generation in the context of de
novo drug design in particular. Rather than generating \emph{any} molecule with a desired
property, we believe that an additional condition on the structural explanation within a
molecule allows a user to generate structures that have the desired property for the
\emph{right reason}.

While the proposed method shows promising results, our experiments also show that particular
combinations of a concept and a property range are much harder to target than others. This may
be the result of a lacking empirical coverage of the joint distribution, or of inherent physical
constraints. In the future, it will be important to investigate ways of bridging these remaining gaps, in
order to enable even more flexible generation and to open up a larger design space.

\ifdefined\submissionbuild
\input{sections/08_experimental}
\fi

\ifdefined\arxivbuild\else\def\showbroaderimpact{}\fi
\ifdefined\showbroaderimpact
\fi

\makeatletter
\if@accepted
\subsubsection*{Author Contributions}
J.T. conceived the project, supervised the initial proof-of-concept study,
designed and implemented the final method and software framework, performed
the computational experiments, and wrote the manuscript. C.B. developed the
original proof-of-concept implementation in the context of a thesis and
proofread the manuscript. P.F. supervised the project, secured funding, and
edited the manuscript.

\subsubsection*{Acknowledgments}
The authors acknowledge support from the state of Baden-W\"urttemberg through
bwHPC. Parts of this work were performed on the HoreKa supercomputer funded by
the Ministry of Science, Research, and the Arts Baden-W\"urttemberg and by the
Federal Ministry of Education and Research. We acknowledge funding by the
German Research Foundation (Deutsche Forschungsgemeinschaft, DFG) within
Priority Programme SPP~2363. This work was supported by funding from the pilot
program Core-Informatics of the Helmholtz Association (HGF).
\ifdefined\submissionbuild

Large language models were used as general-purpose assistive tools during the preparation of this manuscript and
the accompanying software. The authors take full responsibility for all content.

\subsubsection*{Conflicts of Interest}
The authors declare no conflicts of interest.

\subsubsection*{Data Availability Statement}
\availabilitystatement
\else
\ifdefined\arxivbuild
\subsubsection*{Code and Data Availability}
\availabilitystatement
\fi
\fi
\fi
\makeatother

\ifdefined\submissionbuild\bibliographystyle{unsrtnat}\else\bibliographystyle{tmlr}\fi
\bibliography{main}

\fi %

\ifmainpart\else
\label{endofmain}
\ifsipart\else\clearpage\fi
\appendix
\ifsipart
\renewcommand{\thesection}{S\arabic{section}}
\renewcommand{\thefigure}{S\arabic{figure}}
\renewcommand{\thetable}{S\arabic{table}}
\renewcommand{\theequation}{S\arabic{equation}}
\fi
\section{Method Details}
\label{app:method}

\subsection{Concepts and Membership from a Self-Explaining Model}
\label{app:megan}

In this section, we describe in more detail how we obtain the concept vocabulary and the
membership vector $\phivec$ of \Secref{sec:method:megan}, on which the concept adapter is
conditioned. The procedure itself is the method of \citet{teufel2023megan} and
\citet{teufel2024global}, and we restate it here only to the extent that the rest of our
pipeline depends on it.

\paragraph{Explanation Channels and Co-Training.}
\megan{} is an attention-based graph neural network in which every message-passing layer
consists of $K$ parallel attention channels. In addition to the prediction $\hat{y}$, the
forward pass of \Eqref{eq:megan} therefore yields the node importances
$\mV \in [0,1]^{N \times K}$ and the edge importances $\mE \in [0,1]^{|\gE| \times K}$, which
assign one importance value per channel to every node and every edge of the graph. For the
details of the attention mechanism that produces these values, we refer to the original work
\citep{teufel2023megan}.

With this architecture alone, however, there is no mechanism that ensures that the individual
channels develop according to an intended interpretation, such as positive and negative
evidence for the predicted value. For this purpose, the model is trained with an additional
explanation co-training objective. The node importances of each channel are pooled over the
graph, and the pooled values alone are used to solve an approximation of the original
prediction task, for which each channel only uses those training graphs that align with its
intended interpretation. The total training loss then consists of three terms,
\begin{equation}
\gL = \gL_{\mathrm{pred}} + \gamma \, \gL_{\mathrm{exp}} + \beta \, \gL_{\mathrm{sparse}} ,
\label{eq:app-megan-loss}
\end{equation}
where $\gL_{\mathrm{pred}}$ is the prediction loss, $\gL_{\mathrm{exp}}$ the explanation
co-training loss and $\gL_{\mathrm{sparse}}$ an $L_1$ regularization of the node importances
that promotes sparse explanations, while the hyperparameters $\gamma$ and $\beta$ weight the
two additional terms against the prediction.

For a regression task, the interpretation of the channels is defined relative to a reference
value $y_c$, which by default is the mean of the training targets and which splits the training
graphs into two groups. One channel is then trained to collect the substructures that raise the
target value above this reference, and the other channel those that lower it.
For the synthetic graphs of Experiment~1, for example, these are intended to be the red and the
blue motifs respectively. For a classification task, there is one channel for each class.
Consequently, every model in this work has $K = 2$ channels, to which the experiments refer by
their index.

\paragraph{Explanation Embeddings.}
\citet{teufel2024global} define a concept as a cluster in a latent space of explanation
embeddings, which represent only the highlighted substructures of a graph. To obtain such an
embedding, the node embeddings are pooled separately for each channel $k$, with the node
importances of that channel as the pooling weights,
\begin{equation}
\mathbf{z}^{(k)}(G) = \sum_{i=1}^{N} \mV_{i,k} \, \mathbf{h}_i ,
\label{eq:app-megan-embed}
\end{equation}
where $\mathbf{h}_i$ is the embedding that the final message-passing layer of \megan{} assigns
to node $i$, which is unrelated to the states of the frozen generator in \Secref{app:adapter}.
Since a node outside of the explanation of channel $k$ has an importance
close to zero, it barely contributes to this sum. Consequently, $\mathbf{z}^{(k)}$ is mainly
determined by the substructure that the channel highlights, and two graphs that contain the
same motif in different surroundings can receive similar embeddings.

\citet{teufel2024global} furthermore train these embeddings with an additional contrastive
objective, which pulls the embeddings of structurally similar explanations together and pushes
all others apart. As a result, the distances in this latent space reflect the structure of the
highlighted motifs in addition to their effect on the prediction. We believe that this is the
property on which the clustering of the next step relies.

\paragraph{Concept Extraction.}
To extract the concept vocabulary, we compute the explanation embeddings of all graphs that the
model was trained on and cluster them separately for each channel with a density-based
clustering algorithm \citep{campello2013hdbscan}. We choose a density-based method for this
step because it does not require the number of concepts to be determined beforehand and because
it does not assign every embedding to a cluster. A graph whose explanation does not resemble
any recurring motif therefore remains unassigned and is not forced into the nearest cluster.

The clusters of all channels together form the vocabulary of $\nconcepts$ concepts. We will
henceforth refer to the graphs in the dense interior of a cluster as the \emph{core members} of
the corresponding concept, since these are the graphs that the clustering assigns to the
concept without ambiguity. Throughout this work, the core members serve as the reference set of
a concept, both when we describe what a concept represents and when we construct the request of
\Eqref{eq:app-megan-request}. We do not merge or split the clusters after the clustering.
Furthermore, the names by which the figures and tables of this work refer to individual
concepts are our own interpretation of what their core members have in common, and they are not
an output of the procedure itself.

\paragraph{Membership Vector.}
Based on this vocabulary, every graph is assigned a membership vector
$\phivec \in [0,1]^{\nconcepts}$, regardless of whether it is part of the training data or
produced by the generator. Its entry $\phic$ is the distance between the explanation embedding
of the graph and its nearest neighbors among the members of concept $c$, sharpened by a
sparsemax \citep{martins2016sparsemax}. A value of $\phic = 0$ therefore indicates core
membership in the concept, while a value of $\phic = 1$ indicates that the graph is not assigned
to it. We consider a graph to be a member of the concept if $\phic < 0.5$. At this point, it should be
noted that this convention is the opposite of a similarity, which means that a low value
indicates a strong assignment in all tables of this work. Furthermore, the memberships are not
mutually exclusive. A graph can be a member of several concepts at the same time, and it can
also belong to none of them, which is the case if its explanation does not match any recurring
motif of its channel.

The simplest way to request a concept $c$ would be a one-hot vector. Instead, we use the mean
membership vector of the core members of that concept,
\begin{equation}
\phivec^* = \frac{1}{|\gC_c|} \sum_{G \in \gC_c} \phivec(G) ,
\label{eq:app-megan-request}
\end{equation}
where $\gC_c$ is the set of core members of the requested concept $c$. In this way, the request
has the same form as the membership vector of a real member of the concept. We believe that
this is important because the concept adapter is trained on the membership vectors of real
graphs, among which a one-hot vector does not occur.

\paragraph{Where the Model Enters the Pipeline.}
In our pipeline, the same \megan{} model is used for three different purposes. It provides the
condition on which we train the concept adapter, the concept term of the steering energy in
\Eqref{eq:fk-energy}, and the concept match by which we evaluate a generated sample. For the
real-world data, it additionally provides the property that the request names. To score a
generated graph, we first convert its attributes into the input features of \megan{}, and then process the graph with one model forward pass.

However, a possible pitfall of this setup needs to be pointed out. Since the condition, the
energy and the evaluation all originate from the same model, a concept match only states that
this model recognizes its own request in the sample, which does not necessarily mean that the
sample contains the corresponding substructure.
This has to be kept in mind when interpreting the concept match rates of this work. For this
reason, \appref{app:inpainting} additionally evaluates the samples of one comparison by a
substructure pattern that does not involve \megan{} at all.

\subsection{Adapter Architecture and Training}
\label{app:adapter}

In this section, we describe the adapter of \Secref{sec:method:adapter} in full detail. All
adapters in this work share the same architecture and only differ in the condition on which
they are trained. Figure~\ref{fig:adapter-layer} illustrates one transformer layer of the frozen
base together with the two mechanisms that the adapter attaches to its output.

\begin{figure}[t]
\centering
\includegraphics{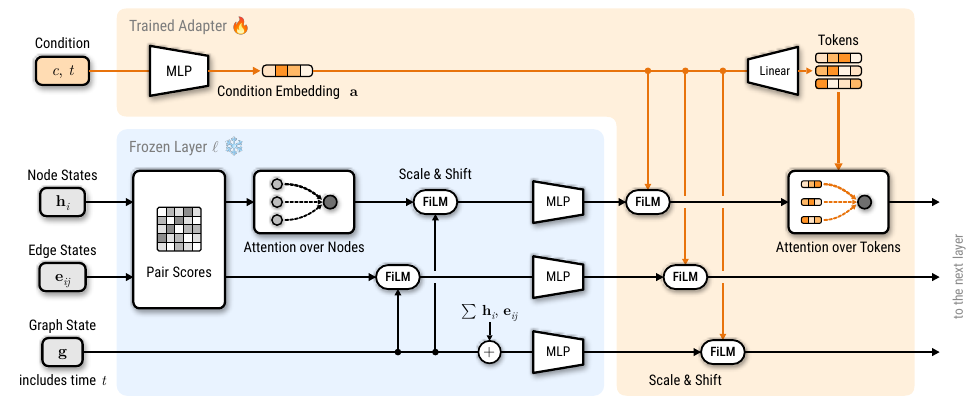}
\caption{\textbf{One layer of the frozen base with the adapter attached.}
Every layer carries a state for each node, each node pair and the whole graph, where the graph
state also holds the flow time. In the frozen layer (blue), scores for all node pairs are computed
from the node and edge states and serve as the weights of an attention over the nodes and as the
update of the edge states, before the graph state modulates both updates by a scale and a shift
\citep[FiLM;][]{perez2018film}. The adapter (orange) attaches the same two mechanisms to the
output of the layer with the condition as their source, a gated FiLM of all three states by the
embedding $\mathbf{a}$ (\Eqref{eq:app-adapter-film}) and an attention of every node over the
condition tokens (\Eqref{eq:app-adapter-xattn}). Residual connections and normalization layers
are omitted.}
\label{fig:adapter-layer}
\end{figure}

\paragraph{Condition Embedding.}
As a first step, the adapter standardizes its condition as $\hat{c} = (c - \mu_c) / \sigma_c$,
where $\mu_c$ and $\sigma_c$ are the mean and the standard deviation of the training labels.
The standardized condition is then mapped to an embedding together with the flow time $t$,
\begin{equation}
\mathbf{a} = \mathrm{MLP}_{\psi}\bigl( [\, \hat{c},\, \boldsymbol{\omega}(\hat{c}),\,
\boldsymbol{\tau}(t) \,] \bigr) \in \R^{d_a} ,
\label{eq:app-adapter-embed}
\end{equation}
where $\boldsymbol{\tau}$ is a sinusoidal time embedding and the MLP consists of two linear
layers with SiLU activations and a layer normalization in between. The terms
$\boldsymbol{\omega}(\hat{c}) = [\sin(2\pi f_k \hat{c}), \cos(2\pi f_k \hat{c})]_{k=1}^{F}$ with
$f_k = 2^{k-2}$ are Fourier features of the condition. We include them because they allow the
embedding to resolve small differences between scalar targets, while nearby targets remain
similar. We use these features for the property adapter only ($F = 3$), whereas the concept
adapter receives the membership vector $\phivec^*$ without them.

\paragraph{Gated Feature Modulation.}
Let $\mathbf{h}^{\ell}_i$, $\mathbf{e}^{\ell}_{ij}$ and $\mathbf{g}^{\ell}$ denote the node,
edge and graph-level hidden states at the output of the $\ell$-th of the $L$ frozen transformer
layers. The first mechanism of the adapter is a gated variant of feature-wise linear modulation
\citep[FiLM;][]{perez2018film}. For this purpose, separate linear heads predict a scale, a shift
and a gate from $\mathbf{a}$ for every layer and for each of the three streams. For the nodes,
these are $(\boldsymbol{\kappa}^{\ell}, \boldsymbol{\delta}^{\ell}, \boldsymbol{\alpha}^{\ell}) =
(W^{\ell}_{\kappa} \mathbf{a}, W^{\ell}_{\delta} \mathbf{a}, W^{\ell}_{\alpha} \mathbf{a})$ with
the biases omitted, and the three states are then updated as gated residuals,
\begin{equation}
\tilde{\mathbf{h}}^{\ell}_i = \mathbf{h}^{\ell}_i + \boldsymbol{\alpha}^{\ell} \odot \bigl(
\boldsymbol{\kappa}^{\ell} \odot \mathbf{h}^{\ell}_i + \boldsymbol{\delta}^{\ell} \bigr) ,
\quad
\tilde{\mathbf{e}}^{\ell}_{ij} = \mathbf{e}^{\ell}_{ij} + \boldsymbol{\alpha}^{\ell}_{E} \odot
\bigl( \boldsymbol{\kappa}^{\ell}_{E} \odot \mathbf{e}^{\ell}_{ij} + \boldsymbol{\delta}^{\ell}_{E}
\bigr) ,
\quad
\tilde{\mathbf{g}}^{\ell} = \mathbf{g}^{\ell} + \boldsymbol{\alpha}^{\ell}_{G} \odot \bigl(
\boldsymbol{\kappa}^{\ell}_{G} \odot \mathbf{g}^{\ell} + \boldsymbol{\delta}^{\ell}_{G} \bigr) .
\label{eq:app-adapter-film}
\end{equation}
Since this modulation acts on the output of each frozen layer, the computation inside the layer
itself remains untouched. Padded nodes and node pairs do not receive an update.

\paragraph{Node-to-Condition Cross-Attention.}
The modulation of \Eqref{eq:app-adapter-film} applies the same diagonal affine map to every
node, so that two nodes with the same state receive the same correction regardless of their
role in the graph. As its second mechanism, the adapter therefore expands the embedding into
$m$ condition tokens,
$[\mathbf{t}_1, \dots, \mathbf{t}_m]^{\top} = \mathrm{reshape}(W_T \mathbf{a}) \in
\R^{m \times d_t}$, and every layer additionally adds a cross-attention residual from the nodes
to these tokens,
\begin{equation}
\mathbf{h}^{\ell+}_i = \tilde{\mathbf{h}}^{\ell}_i + W^{\ell}_O \sum_{k=1}^{m}
\operatorname*{softmax}_k \Biggl( \frac{\bigl( W^{\ell}_Q \operatorname{LN}(
\tilde{\mathbf{h}}^{\ell}_i ) \bigr)^{\!\top} W^{\ell}_K \mathbf{t}_k}{\sqrt{d}} \Biggr)
W^{\ell}_V \mathbf{t}_k ,
\label{eq:app-adapter-xattn}
\end{equation}
which is written here for a single attention head of width $d$, and where $\operatorname{LN}$
is a layer normalization on the query side. In this mechanism, the query is formed from the
state of the node itself, which by the middle layers of the network reflects the neighborhood
of that node. Different nodes can therefore select different tokens and receive qualitatively
different corrections. The edge and graph-level states only receive the modulation of
\Eqref{eq:app-adapter-film}, and $\mathbf{h}^{\ell+}_i$, $\tilde{\mathbf{e}}^{\ell}_{ij}$ and
$\tilde{\mathbf{g}}^{\ell}$ are subsequently passed to layer $\ell + 1$.

\paragraph{Initialization and Training.}
We initialize the gate heads $W^{\ell}_{\alpha}$ of all three streams and the output
projections $W^{\ell}_O$ to zero. Consequently, both residuals of \Eqref{eq:app-adapter-film}
and \Eqref{eq:app-adapter-xattn} vanish, and an untrained adapter reproduces the frozen base
exactly. We then train each adapter on its labeled dataset with the denoising objective of the
base,
\begin{equation}
\mathcal{L}(\psi) = \mathbb{E}_{t,\,(G, c),\, z_t} \Bigl[ - \sum_{i} \log
p_{\theta,\psi}(x_i \mid z_t, c) - \lambda \sum_{i < j} \log
p_{\theta,\psi}(u_{ij} \mid z_t, c) \Bigr] ,
\label{eq:app-adapter-loss}
\end{equation}
where $z_t$ is the graph noised to time $t$ as in \defog{} and $\lambda$ is the edge weight of
the base objective, while only the adapter parameters $\psi$ receive gradients. We do not use
condition dropout during this training, since the unconditional branch that \Eqref{eq:cfg} and
\Eqref{eq:compose} require is the frozen base itself.

\paragraph{Settings.}
All adapters use an embedding width of $d_a = 256$, $m = 64$ condition tokens of width
$d_t = 128$ and $16$ attention heads. They are trained with AdamW at a learning rate of
$4 \times 10^{-4}$ and a weight decay of $10^{-5}$. The modulation heads cover all $L$ layers
and all three streams of the base.

\section{Finding Concept Prototypes}
\label{app:prototypes}

The main paper presents a method which can be used to steer a generative model for graphs toward graphs that not only match a desired property, but also
match that property for a specific structural reason, as represented by a \emph{concept explanation}. This section introduces an additional use case that
can be derived from this core functionality, but lies slightly outside the central scope of the paper. It shows how the proposed graph generation method
can be used to create prototypes, which make it easier to identify the core motifs that underlie concept explanations.

\subsection{Graph Concept Explanations}
\label{app:prototypes:concepts}

To motivate this use case properly, it makes sense to reintroduce what graph concept explanations are and how they are represented in this work
(\Secref{sec:method:megan}). Generally speaking, a concept explanation is a global explanation of the behavior of a model, which associates small and
recurring input fragments, the ``concepts'', with a specific tendency in the prediction behavior of the model.

This can be illustrated in the image processing domain, where we consider a hypothetical model that classifies photographs by the type of sport that they
depict, such as basketball, football, or golf. Each photograph may show a completely different global scene, but the photographs still contain recurring
details or image fragments, which the model likely uses as indicators for certain classes. For instance, one small image fragment that may appear in many
photographs is a basketball, which the model likely correlates strongly with the basketball class. Another example is patches of greenery such as trees or
grass, which the model may not correlate with any specific sport, but which it likely correlates more highly with outdoor sports such as golf. The set of
all such image fragments makes up the ``concept'' as it is presented to the user, whereas the concept itself is closer to an abstract idea. In this case,
the user sees multiple small image fragments of basketballs, for instance, but the cognitive jump from these fragments to the abstract idea of a basketball
is still left to the user. Making this connection between a set of fragments and the underlying shared idea is usually not an issue for simple problems
such as the hypothetical sports classification, but it becomes substantially more challenging for harder tasks.

A harder example of concept explanations can be found in molecular property prediction, where a model predicts properties of molecular graphs instead of
classifying images, and the concept explanations are subgraph fragments instead of image fragments. As a concept, the user is therefore presented with a
set of molecules and their local attributional explanations, which highlight different subgraphs. Connecting this set of molecules to the more
fundamental, abstract ``idea'' of a functional group generally requires a certain level of chemical expertise. Consequently, it would be desirable if this
step could be automated or at least simplified to some extent.

A \emph{prototype} is a type of explanation that is related to concepts, in the sense that it also represents a global explanation of model behavior. In
contrast to a concept, a prototype explanation presents only a single, most prototypical instance (or a very limited set of them) to the user as a
representation of the underlying abstract idea. As such, it may be easier to understand, but at the very least it presents an additional interpretation
aid when shown alongside the concept set. The following sections therefore explore how the conditional graph generation pipeline presented in this paper
can help with finding concept prototypes for an existing concept set.

\subsection{Finding Concept Prototypes with Conditional Graph Generation}
\label{app:prototypes:method}

The conditional graph generation pipeline presented in \Secref{sec:method:adapter} and \appref{app:adapter} can generate graphs with the additional
condition that the generated graph should be a member of a requested concept $c$, which is expressed through its target membership $\phivec^*_c$. It is
this property that can be used to generate concept prototypes. In the following, we define a concept prototype graph as the smallest possible graph that
is still associated with a given concept. In other words, it is a graph that consists almost purely of nodes that are relevant for the concept and
contains a minimal amount of irrelevant, noisy nodes.

One technical property of generating graphs with discrete flow matching is that the number of nodes has to be set prior to generation and cannot be
changed during the denoising process. The generated graph will have exactly the number of nodes configured at the beginning. One practical consequence of
this is, for example, that matching the size distribution of a dataset requires modeling this distribution and sampling from it explicitly for the
generation process. However, being able to control the number of nodes of a generated graph exactly also has advantages, for instance if there is a hard
constraint on the graph size. It is also this property that can be used to find prototypes as defined above.

To find concept prototype graphs, we start by generating a fixed number of graphs (for instance $128$) at the average size that is associated with the
concept. We then decrease the size iteratively in a sweep and generate another set of graphs for each size. For each set of generated graphs, we use the
\megan{} model to determine the share of graphs that match the concept, which we call the hit rate of the set. Since we assume that the concept
conditioning works, this rate will likely be high at first ($80\%$ for instance). However, since every concept represents some kind of motif, which has a
size greater than zero, the hit rate approaches zero as the sweep approaches a graph size of zero.

We therefore define two possible methods for determining concept prototype graphs from this sweep. The cautious method applies an elbow criterion to the
hit rate as a function of the graph size. At the size of the ``elbow'' of the decrease in hit rate, we choose the graph that is generated most often among
the hits of that size as the concept prototype. We choose the most frequent hit so that the prototype is a graph that the generator reaches reliably, and
ties are decided by the strongest membership. A more aggressive method could simply look for the smallest single graph among all generated graphs that satisfies a minimal threshold of concept
membership. We believe that the aggressive method finds smaller graphs, but at the risk of selecting erroneous, out-of-distribution samples. The
cautious method, on the contrary, should select prototypes that are still well within the distribution, but that may not be entirely minimal.

\subsection{Experiments}
\label{app:prototypes:experiments}

To investigate this method, we conduct an experiment on the mutagenicity dataset of Experiment~4 with the goal of identifying a minimally representative
prototype for each concept, which simplifies the process of finding the abstract idea behind a cluster of molecules. We use the frozen base, the \megan{}
model and the concept adapter of that experiment, with the adapter alone ($w = 1$) and without steering, and we apply the cautious method. For each
concept $c$, we sweep the number of nodes $n$ from the median size of the real core members of the concept down to $3$ and draw $N = 128$ graphs for each
size. A graph is a hit if it is valid, connected, plausible and has a membership $\phic < 0.5$ in $c$, and the hit rate $h_c(n)$ is the share of hits among
the graphs of size $n$. We call a graph plausible if it contains neither two fused three-membered rings nor a hydrogen on a sulfur atom with more than
three neighbors.

We define the plateau $P_c$ as the median hit rate over the largest third of the swept sizes and the elbow as the smallest size at which the hit rate still
reaches half of it,
\begin{equation}
n^*_c = \min \left\{ n : h_c(n) \ge P_c / 2 \right\} .
\label{eq:elbow}
\end{equation}
The prototype is the graph that occurs most often among the hits at the elbow, with ties decided by the smaller $\phic$, and a concept with a plateau below
$0.2$ gets no prototype. As two checks that do not rely on the adapter or on \megan{} as the grader, we also draw $256$ graphs per size from the frozen base alone and
count the graphs that contain the defining substructure of the concept. \appref{app:protocol:prototypes} describes the full protocol.

\begin{figure}[!tp]
\centering
\includegraphics{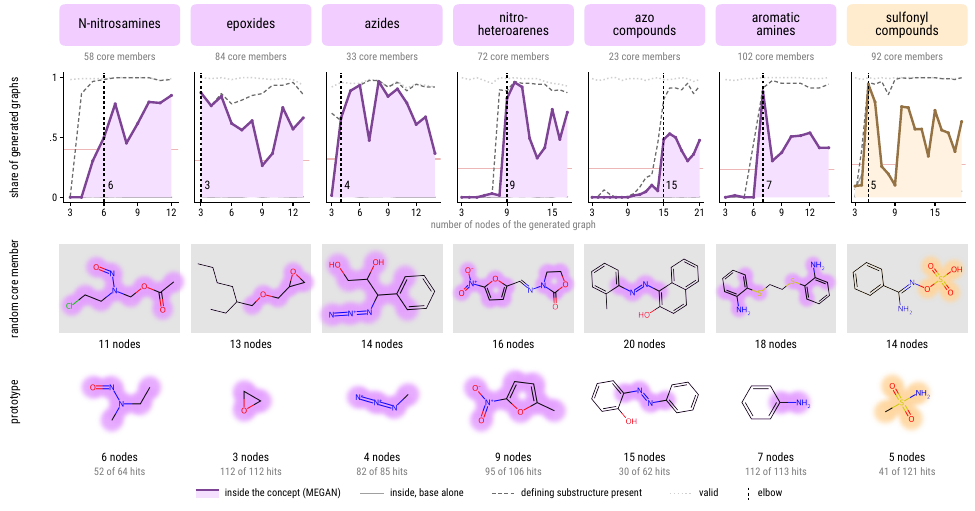}
\caption{\textbf{Hit rate by size, a random core member and the prototype of the seven named concepts of the real-world experiment.} Each column is a
concept of \megan{}, named as in Figure~\ref{fig:e5}, swept from the median size of its real core members down to $3$ nodes. Top: the share of the $128$
generated graphs that \megan{} assigns to the concept (thick line), that contain its defining substructure (dashed), that the frozen base alone assigns to
it (grey) and that are valid (dotted); the thin red line marks half of the plateau and the vertical dashed line the elbow. Below: a random core member of
the concept (grey), drawn among the core members that contain the defining substructure and have a size between the quartiles of all core members, and the
prototype at the elbow, with its number among the hits of that size. Highlighted nodes are the explanation of \megan{} in the channel of the concept
(importance above $0.15$), purple for channel~1 and orange for channel~0.}
\label{fig:prototypes}
\end{figure}

Figure~\ref{fig:prototypes} shows the hit rate by size, a random core member and the prototype for the seven named concepts of the real-world experiment.
Over all $29$ concepts of the mutagenicity \megan{} model, the sweep finds a prototype for $22$, and the other $7$ have a plateau below $0.2$. We find that
this method is able to find small and reliable representations of the underlying motifs of a concept, even for concepts of very different sizes, such as
the azides, whose prototype has $4$ nodes, and the nitro-heteroarenes, whose prototype has $9$. In both cases, the prototype is the graph that the adapter
generates most often among the hits at the elbow, with $82$ of $85$ hits for the azides and $95$ of $106$ for the nitro-heteroarenes. The prototypes of the
seven named concepts have between $3$ and $15$ nodes, whereas the real core members of these concepts have a median of $12$ to $21$ nodes.

Additionally, we observe that this method seems to be able to take important environment effects into account, such as for the azo compounds. The members
of this concept are characterized by an azo bridge specifically between two aromatic ring systems. While the explanation mask for this concept highlights
only the azo bridge and the two ring atoms attached to it, the prototype that we find for it does correctly retain also both ring systems outside of the
explanation mask, which are apparently important for the concept membership assignment.
\section{Steerability Ablations}
\label{app:fk_adapter}

This section contains two additional analyses of the joint conditioning on the synthetic graphs
of Experiment~2. The first one investigates how the results depend on the guidance weights of
the adapters and on the weighting of the steering energy (\Secref{app:fk_adapter:sweep}), and
the second one compares the adapter-based approach with the original conditioning mechanism of
\defog{} (\Secref{app:fk_adapter:native}).

\subsection{Guidance Weights and Steering Strength}
\label{app:fk_adapter:sweep}

\begin{figure}[!tp]
\centering
\includegraphics{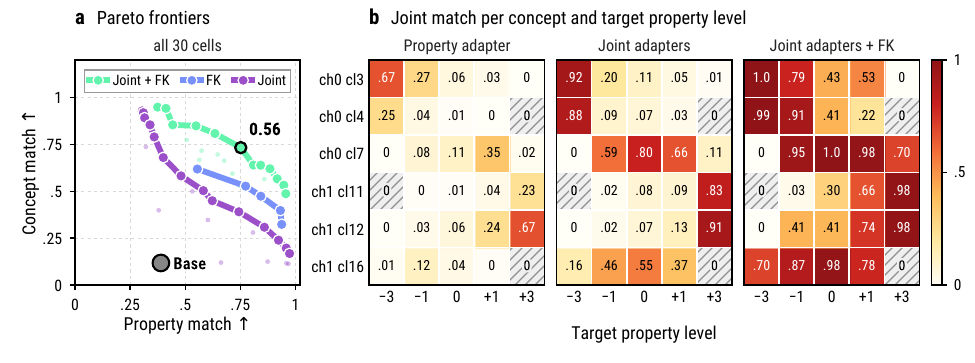}
\caption{\textbf{Joint conditioning across adapter weights and steering strengths.}
\textbf{a}~Pareto frontiers of property match and concept match over all $30$
target cells. Each line connects the settings that no other setting of the same
method beats on both axes: the joint adapters (Joint, purple) over the guidance
weights of both adapters, steering alone (FK, blue) over $\lambda$, and the joint
adapters with steering (Joint + FK, green, ours) over four weight pairs and
$\lambda$. Faint dots show
all $49$ settings tried. The ringed point marks the highest joint match of our
method ($0.56$), and the gray dot is the unconditioned base. \textbf{b}~Joint
match per concept and target property level for the property adapter alone, the
composed adapters ($w_c = w_p = 1$) and the composed adapters with steering
($\lambda = 1$), as in Table~\ref{tab:e2}. Each cell is the fraction of $128$
generated graphs that satisfy both conditions, and the three heatmaps share one
color scale. Hatched cells have no training graph with that concept at that
level, and every method scores zero there.}
\label{fig:e2}
\end{figure}

The main text reports every arm of Experiment~2 at a single setting (Table~\ref{tab:e2}). In
this experiment, we investigate whether the comparison between the arms depends on this choice.
Our method has two kinds of free parameters, namely the guidance weights $(w_c, w_p)$ of the
concept and the property adapter in \Eqref{eq:compose} and the weight $\lambda$ that the
steering energy of \Eqref{eq:fk-energy} assigns to the property. For this purpose, we repeat
the evaluation on the same $30$ target cells with $128$ graphs per cell for three groups of
settings. For the composed adapters, we vary both guidance weights over $\{0.5, 1, 2, 3\}$,
which results in $16$ pairs. For Feynman--Kac steering of the base alone, we vary $\lambda$
over $\{0.25, 0.5, 1, 2, 4\}$, and for the composed adapters with steering, we combine the same
values of $\lambda$ with the four weight pairs $(w_c, w_p) \in \{1, 2\}^2$. Together with the
two individual adapters at four weights each, this amounts to $49$ settings
(Figure~\ref{fig:e2}a).

The results show that the guidance weights mainly trade one condition against the other.
Increasing the weight of the property adapter raises the property match of the composed adapters
and lowers their concept match, and vice versa, so that the $16$ pairs cover property matches
between $31\%$ and $97\%$ and concept matches between $17\%$ and $92\%$. The joint match,
however, remains between $16\%$ and $32\%$ for all pairs. This indicates that the limited
joint match of the composed adapters is not caused by the choice of the weights. We observe the
same for the weight of the steering energy, since the joint match stays between $30\%$ and
$39\%$ for steering alone and between $36\%$ and $56\%$ for the composed adapters with steering.

Consequently, the ordering of the arms that the main text reports is largely independent of
these settings. All $20$ settings of the composed adapters with steering achieve a higher joint
match than every weight pair of the composed adapters alone, and $19$ of them also exceed the
best setting of steering alone. The highest joint match of $56\%$ is reached at the weights
$(w_c, w_p) = (1, 1)$ with $\lambda = 1$, which is the configuration of Table~\ref{tab:e2}. The sweep uses $128$ graphs per
cell, whereas Table~\ref{tab:e2} was repeated with $1024$ graphs per cell afterwards, and the same configuration reaches
$54\%$ there, which we attribute to the smaller sample of the sweep.
Figure~\ref{fig:e2}b additionally resolves the joint match by concept and property level.
Steering improves the joint match in the native as well as in the off-native cells (from $64\%$
to $93\%$ and from $11\%$ to $44\%$). In the three cells for which the training data contain
no graph with the requested combination, on the other hand, none of the three arms shown
produces a match, and the joint match does not exceed $3\%$ in any setting of the sweep.

However, two limitations of this result have to be noted. Firstly, the weights $(1, 1)$ of the
composed adapters in the main text were fixed among three pairs before this sweep was conducted
(\appref{app:protocol}), and they are not the best pair of the larger grid. The pair
$(2, 2)$ achieves a joint match of $32\%$ compared to $27\%$, which does not change the ordering
of the arms. Secondly, steering reduces the uniqueness of the samples to between $75\%$ and
$89\%$ in all of its settings, while the samples of the adapters alone remain unique.

\subsection{Comparison with the Original Conditioning of \defog{}}
\label{app:fk_adapter:native}

The adapters of this work replace the conditioning mechanism of the original \defog{} model
(\Secref{sec:method:defog}). In that mechanism, the condition is an input of the denoising
network, and the whole network is trained on the labeled data with classifier-free guidance. A
natural question is how much conditioning quality the adapter-based approach gives up in
exchange for its modularity. In this experiment, we therefore compare the composed adapters with
a \defog{} model that uses the original mechanism for the joint condition. For this purpose, we
train a model with the architecture and the training configuration of the frozen base end to
end on the same synthetic graphs. Its condition is the concatenation of the standardized
property value and the membership vector $\phivec$ of each graph. We will henceforth refer to
this model as the end-to-end model. We then evaluate it on the $30$ target cells of
Experiment~2 with the same requests and the same metrics. We sample it at a guidance weight of
$\gamma = 1$ and without Feynman--Kac steering, which corresponds to the composed adapters at
the weights $(w_c, w_p) = (1, 1)$.

\begin{table}[t]
\centering
\small
\setlength{\tabcolsep}{3.5pt}
\caption{Comparison of the end-to-end model with the composed adapters on the synthetic graphs
for the $30$ target cells and the metrics of Table~\ref{tab:e2}. The end-to-end model receives
both conditions through the original conditioning mechanism of \defog{} and is sampled at a
guidance weight of $1$. The two rows of the composed adapters are repeated from
Table~\ref{tab:e2}, and the last row additionally uses Feynman--Kac steering.}
\label{tab:native}
\begin{tabular*}{\textwidth}{l @{\extracolsep{\fill}} c @{\extracolsep{0pt}\hspace{2\tabcolsep}} cc cc cc r}
\toprule
 & \multicolumn{3}{c}{Match $\uparrow$} & \multicolumn{2}{c}{Joint by cell $\uparrow$} & & & \\
\cmidrule(lr){2-4}\cmidrule(lr){5-6}
Model & Prop. & Concept & Joint & Native & Off-nat. & Unique $\uparrow$ & Valid $\uparrow$
      & Cost $\downarrow$ \\
\midrule
End-to-end model             & 0.744 & 0.456 & 0.359 & 0.793 & 0.165 & 1.000 & 0.988 & 3.0 \\
Both adapters                & 0.541 & 0.500 & 0.276 & 0.646 & 0.107 & 1.000 & 0.993 & 7.2 \\
\midrule
Both adapters $+$ FK         & 0.732 & 0.741 & 0.544 & 0.932 & 0.412 & 0.817 & 0.999 & 2.2 \\
\bottomrule
\end{tabular*}
\end{table}

The results show that the end-to-end model achieves a higher joint match than the composed
adapters, with $36\%$ compared to $28\%$ (Table~\ref{tab:native}). The difference is found in
the property condition, for which the end-to-end model reaches a match of $74\%$ compared to
$54\%$, while its concept match is slightly lower ($46\%$ compared to $50\%$). The advantage is
consistent across the target cells, as the end-to-end model has the higher joint match in $23$
of the $27$ supported cells. In the three cells without training data, neither approach is
successful, as the end-to-end model produces $42$ joint matches among $3072$ graphs and the
composed adapters $2$. The advantage does not depend on the guidance weights of the adapters
either, since the best of the $16$ weight pairs of \Secref{app:fk_adapter:sweep}, sampled with $128$ graphs per cell,
reaches a joint match of $32\%$.

This indicates that the modularity of the adapters comes at the cost of about a quarter of the
joint match that the end-to-end model achieves. We believe that this is a consequence of the
separate training. The end-to-end model observes both conditions together for every training
graph, whereas the two adapters are trained independently of each other and are only combined
at sampling time. In return, the end-to-end model has to be trained from scratch for this
particular pair of conditions, while each adapter is trained on the frozen base and can be
combined with others. Moreover, Feynman--Kac steering more than compensates for the difference
in this experiment. The composed adapters with steering reach a joint match of $54\%$ and
exceed the end-to-end model in $25$ of the $27$ supported cells.

\section{Relationship to Inpainting}
\label{app:inpainting}

The main paper proposes a method to condition a graph generation model to not only match a desired property, but also to contain a certain explanatory
subgraph motif. The process of matching a certain subgraph motif to be contained in the generated graph is adjacent to the process of \emph{inpainting}
and could consequently be compared to the latter. However, in this section we argue that concept conditioning and inpainting have two different objectives
and also have different technical advantages and disadvantages, which prevent the one from being substituted with the other.

\subsection{Generative Inpainting}
\label{app:inpainting:generative}

The process known as \emph{inpainting} conditions an iterative generative model to contain a specific substructure, be it a part of an image or a subgraph
motif, while generating only the remaining parts of the instance. This process is a passive property of iterative models and can be used for
structure-conditional generation without explicitly training the model for it. For score-based models, \citet{song2021score} show that an unconditional
model can impute a partial sample if, at every step of sampling, the known part is replaced by a draw from its own perturbation at the current noise level,
while the model generates the unknown part around it. Since the known part is perturbed without regard to what has been generated, the two parts can
disagree at their boundary, and the replacement approximates the conditional distribution of the generated part with an error that a more expressive model
does not remove \citep{trippe2023diffusion}. \citet{lugmayr2022repaint} address the disagreement in RePaint by resampling, in which the sampler repeatedly
jumps back a fixed number of steps by adding noise to its current estimate and denoises again, so that the generated part can adapt to the known part, and
they find that the benefit saturates at about ten resamplings. Molecular models adopt this recipe to design around a given part of a molecule:
\citet{schneuing2024structure} fix a part of a ligand and let a pretrained diffusion model generate the rest, with resampling, for scaffold hopping and
elaboration and for growing, merging and linking fragments, and \citet{runcie2023silvr} condition a pretrained equivariant diffusion model on fragments in
a related way, both without retraining, whereas models for scaffold-conditioned generation are otherwise trained for the task
\citep{lim2020scaffold, imrie2020delinker}.

\subsection{Two Different Objectives}
\label{app:inpainting:objectives}

As described above, inpainting introduces a certain substructure into a generated graph by repeatedly replacing the corresponding nodes of the graph with
the desired substructure at intermediate steps of the iterative generation. Additional resampling steps can be used to avoid generation artifacts due to
the mismatch at the boundary between the desired substructure and the freely generated part. This methodology is applied to the generative process only at
inference time and therefore does not require any additional training. Concept conditioning, on the other hand, requires training the generator model (or
an adapter module) conditionally on known concept memberships, which have to be derived beforehand from an additional explanation model or method. Once
active at inference time, concept conditioning aims at a similar objective, which is to promote specific explanatory substructures to appear in the
generated instances.

At first glance, one could regard both methods as different approaches toward the same goal. However, we argue that they serve slightly different purposes
and have different use cases. Most importantly, the explanation process that precedes concept conditioning performs an \emph{informed filtering} of the
motifs down to those motifs that have actually been found to be relevant to the underlying task. To use inpainting, on the contrary, a user would have to
come up with the relevant motifs entirely by themselves and therefore relies on domain knowledge to identify motifs that are relevant for the underlying
task. Another important difference is reflected in how both methods frame the structural condition. Inpainting creates a hard condition for exact nodes,
while concept conditioning achieves the condition over a learned, soft concept membership distribution. This becomes relevant for concepts that are not
hard motifs, but rather vague, overarching ideas. One concept for the prediction of water solubility could be summarized as ``alkyl chains'' (long chains
of carbons), for instance (\Figref{fig:inpainting-regions}). The concept cluster contains molecules with chains of various lengths, whereas in inpainting
one would have to decide on one particular number of carbon atoms. For this particular concept, inpainting would face a different issue as well. While it
guarantees that the given atoms of a motif do not have any additional bonds between themselves, it does not restrict the possible connections to the
remaining atoms. This means that rather than keeping the carbons fixed as a ``chain'', they could be co-opted into a different structure altogether, such
as a ring.

\begin{figure}[t]
\centering
\includegraphics{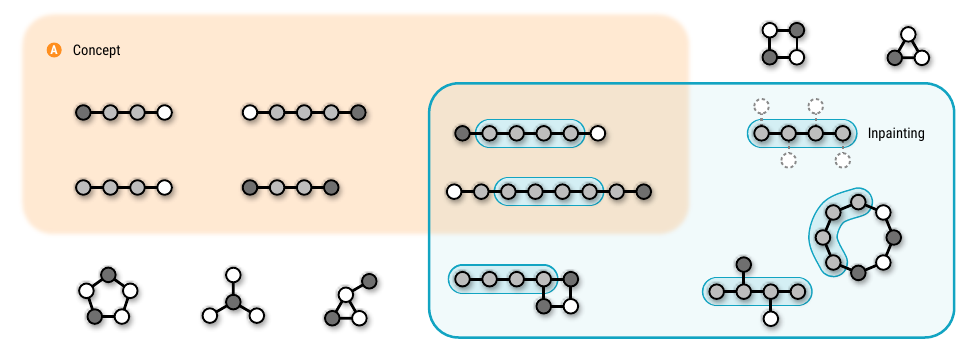}
\caption{\textbf{Visual illustration of the difference between concept conditioning and inpainting.}
In a hypothetical scenario, the underlying structure of a property generally requires a ``chain of gray
nodes''. Concept conditioning (orange) can cover this request diffusely by mapping to different sizes of
gray-node chains. In inpainting (cyan), one has to decide on one static collection of nodes to keep fixed
during the generation, therefore having to decide on a single length of the chain. Moreover, inpainting makes
no additional assumptions about the rest of the graph. This could lead to graphs which contain the requested
gray nodes but use them in patterns other than the desired chain, such as closing them into a ring.}
\label{fig:inpainting-regions}
\end{figure}

Apart from the conceptual differences, inpainting and concept conditioning also have different technical advantages and disadvantages. As described above,
inpainting requires no modification of the training, but it requires up to about ten times more inference time per sample in the case of RePaint
\citep{lugmayr2022repaint}. Concept conditioning, on the other hand, requires a special training, but either no additional inference time, if the model is
trained end to end, or $J + 1$ evaluations of the denoising network per step for $J$ adapters (\Secref{sec:method:adapter}). This is two evaluations for the
concept adapter alone. Concept conditioning does not fully guarantee the occurrence of the desired substructure, but is more likely to generate the correct local
environment around it, as the experiments of \Secref{app:inpainting:experiments} indicate. Inpainting, on the other hand, guarantees the desired nodes to be
present, but could co-opt them into a different structure altogether.

\subsection{Implementation}
\label{app:inpainting:implementation}

We also implement RePaint-style inpainting for graphs, so that it can be used in conjunction with the property conditioning proposed in the main paper and
with the concept steering of \Secref{sec:method:fk}. The recipe carries over to discrete flow models. The noising process of \defog{}
(\Secref{sec:method:defog}) keeps the value of every node and node pair with probability $t$ and draws it from the prior $\pi$ otherwise. A known value
$x_1$ can therefore be brought to the noise level of the rest of the graph at any time $t$ in closed form,
\begin{equation}
x_t \sim t\, \delta_{x_1} + (1 - t)\, \pi .
\label{eq:inpainting-noise}
\end{equation}
The purpose is to keep the input of the denoising network on the distribution that it was trained on. The sampler of our \defog{} implementation provides
this procedure for graphs, written after RePaint, and we use it unchanged with one addition, a graph size for every sample.

The known part is a core of $k$ nodes, which occupy the first $k$ places of every graph. Inpainting fixes the classes of its nodes and of all pairs among
them, so that bonded pairs stay bonded and non-bonded pairs stay non-bonded. It generates every other node and every pair that involves one, including
the pairs between a core node and a new node, so that a bond from the group to its surroundings, or the absence of one, is left to the sampler. For a group
of a molecule, the core is its heavy-atom graph in the vocabulary of the base, which has no aromatic bond class, so that an aromatic ring is written with
alternating single and double bonds, as in the training data. The cores of \Secref{app:inpainting:experiments} have $7$ nodes for chlorophenyl,
bromophenyl and the sulfolane ring, and $5$ for thiophene. The sampler takes one size for a whole batch, whereas every other arm of the paper draws one
graph size per sample from the size distribution of the base, so that the size cannot depend on the target. We therefore draw the size of every sample from
that distribution, restricted to sizes that leave room for at least one free node.

Sampling runs the $500$ steps of the other arms. Before every step, at the current time $t$, the core is overwritten by a draw from
\Eqref{eq:inpainting-noise} with its fixed values as $x_1$, so that it is as noisy as the rest of the graph, and after the last step the exact core is
installed. We add resampling after RePaint, which we call time travel to keep it apart from the resampling of particles in Feynman--Kac steering
(\Secref{sec:method:fk}). After every $50$ steps, the sampler jumps back $50$ steps and denoises them again, four times in succession. At each jump it
draws the whole graph from \Eqref{eq:inpainting-noise} at the earlier time with its current values as $x_1$, since a discrete flow has no closed-form
backward kernel, pins the core at that time and repeats the $50$ steps. This costs $4.6$ times the forward passes of plain sampling, which is lighter than
the ten resamplings with a jump of ten steps in the experiments of RePaint, which would cost $10.8$ times, and we did not tune the schedule.

The property adapter conditions the denoising network on the target value as in the other arms, at the weight $w_p = 1$ that a pilot of $32$ graphs per
cell selected before the main run. The steered arm applies the same two operations before every step and after the last inside the Feynman--Kac sampler
of \Secref{sec:method:fk}, with the $16$ particles and the energy of the other steered arms, but without time travel, which our implementation does not
offer inside this sampler. This energy contains the membership term of the requested concept, so that steering can move the surroundings of the frozen
group toward the concept, which the frozen group alone cannot do. We check in the graph of every sample that the first $k$ nodes and all pairs among them
are exactly the core, which holds for all $30{,}720$ graphs of the three arms that freeze a group. After decoding to a molecule, however, the free bonds
can make the core part of a larger ring system in which its atoms are aromatic, and the pattern of the group then fails although the graph kept the core.

\subsection{Computational Experiments}
\label{app:inpainting:experiments}

Despite the conceptual differences, we nonetheless conduct some basic computational experiments to compare inpainting and concept conditioning on those
concepts that do represent one strict motif explicitly. We compare the two on the four such concepts of Experiment~3, namely chlorophenyl, bromophenyl, the
sulfolane ring and thiophene, at the five $\log P$ levels, which makes $20$ target cells with $512$ generated graphs per arm and cell. Inpainting freezes
the group and is combined with the property adapter, the concept adapters together with the property adapter are the reference, and each of the two is run
without and with Feynman--Kac steering. We add inpainting alone as a control, which reaches the requested $\log P$ in only $0.180$ of its graphs and is the
reason why we combine inpainting with the property adapter. We grade every molecule twice (Table~\ref{tab:inpaint-group}), by the pattern of the group,
which is the request that inpainting is built to satisfy, and by the membership $\phic < 0.5$, which is the request that the adapters are trained for.

\begin{table}[!htp]
\centering
\small
\setlength{\tabcolsep}{3.2pt}
\caption{Inpainting and the concept adapters on molecular graphs for $20$ target cells of four
concepts that a single group names and five $\log P$ levels, with $512$ generated graphs per
cell, graded by the pattern of the group and by core membership in the requested concept. Under
Group pattern, a match is a valid molecule that contains the pattern of the group, and under
\megan{} concept, it is a valid molecule with $\phic < 0.5$ for the requested concept. A joint
match also has a Crippen $\log P$ within $0.5$ of the target, and Distinct counts each matching
molecule once per cell. Every column except Time is a fraction of all
generated graphs, averaged over the cells, and Time is the sampling time per graph in seconds.
Inpainting freezes the group during sampling with time travel, which the steered arm lacks, and
the control has no property adapter and is not highlighted. The best result per column and
section is highlighted in bold. The half-width of the $95\%$ bootstrap interval of every mean,
over the graphs of one sampling run per arm, is up to about $0.025$.}
\label{tab:inpaint-group}
\begin{tabular}{l cc ccc ccc r}
\toprule
 & & & \multicolumn{3}{c}{Group pattern $\uparrow$} & \multicolumn{3}{c}{\megan{} concept $\uparrow$} & \\
\cmidrule(lr){4-6}\cmidrule(lr){7-9}
Model & Valid $\uparrow$ & Prop. $\uparrow$ & Match & Joint & Distinct & Match & Joint & Distinct & Time $\downarrow$ \\
\midrule
\multicolumn{10}{l}{Without steering} \\
Both adapters                      & \textbf{0.940} & \textbf{0.709} & 0.721          & 0.551          & 0.551          & \textbf{0.340} & \textbf{0.254} & \textbf{0.254} & \textbf{1.14} \\
Inpainting $+$ property adapter    & 0.892          & 0.682          & \textbf{0.867} & \textbf{0.664} & \textbf{0.663} & 0.251          & 0.193          & 0.192          & 3.52 \\
Inpainting alone (control)         & 0.947          & 0.180          & 0.933          & 0.178          & 0.178          & 0.267          & 0.052          & 0.052          & 1.63 \\
\midrule
\multicolumn{10}{l}{With steering} \\
Both adapters                      & \textbf{0.993} & \textbf{0.945} & 0.880          & 0.839          & \textbf{0.273} & \textbf{0.822} & \textbf{0.784} & \textbf{0.244} & 1.21 \\
Inpainting $+$ property adapter    & 0.932          & 0.887          & \textbf{0.919} & \textbf{0.878} & 0.264          & 0.746          & 0.720          & 0.203          & \textbf{0.91} \\
\bottomrule
\end{tabular}
\end{table}

By the pattern, inpainting works as intended and is ahead of the adapters: it places the group in $97\%$ of the molecules at the requested $\log P$ against
$78\%$ for the adapters. Its joint match is higher by $0.113$ without steering ($0.664$ against $0.551$) and by $0.039$ with steering ($0.878$ against
$0.839$; $95\%$ interval $0.017$ to $0.062$). By the membership, the order reverses, with a joint match of $0.193$ against $0.254$ without and $0.720$
against $0.784$ with steering, and the adapters reach the requested concept more often although they contain the group less often, which we believe
reflects that the group is only part of what the concept names.

However, neither grading is neutral. The pattern measures what the clamp guarantees by construction, and the membership measures what the adapters are
trained for and what the steering energy of both families consists of, so that both steered arms move toward the concept and not toward the group. We have
not tested a steering toward the group, which may change the two steered rows. The lead of inpainting under the pattern also shrinks under steering, since
it rests on the sulfolane ring, where inpainting is ahead by $0.208$, whereas on the other three concepts the two arms are level or the adapters are ahead.
If every molecule is counted once, the two arms are level ($0.264$ against $0.273$). Inpainting with time travel also takes $3.1$ times as long per
graph as the adapters without steering ($3.52$ against $1.14$ seconds). Finally, this version of inpainting is neither exact nor tuned, so that we read the
comparison as one against a reasonable version of inpainting and not as a bound on what inpainting can reach. The comparison also covers only concepts that
a single group names, so that it says nothing about concepts that are families of structures, such as the alkyl chains of
\Secref{app:inpainting:objectives}.
\section{Nitrenium-Ion Stability of the Aromatic-Amine Example}
\label{app:nitrenium}

This section tests the chemical reading of the aromatic-amine example of Experiment~4 (the sixth column of Figure~\ref{fig:e5}) with
quantum-chemical calculations. In the molecule generated as mutagenic, the free amine sits across the ring from an acetamido group,
whereas the molecule generated as non-mutagenic places the amine next to a chlorine atom. We read this contrast through the activation of
aromatic amines, which are converted by $N$-hydroxylation and further steps into a nitrenium ion that reacts with DNA
\citep{benigni2011mechanisms}. Groups that stabilize this ion should therefore make an amine more likely to be mutagenic.
\citet{ford1992nitrenium} relate the relative stabilities of the nitrenium ions of polycyclic aromatic amines to their mutagenicity.
\citet{bentzien2010insilico} predict the Ames activity of primary aromatic amines from the calculated stability of these ions, and
\citet{furukawa2022nitrenium} and \citet{muto2024nitrenium} use the same quantity in local models of the mutagenicity of such amines.
Based on this literature, we calculate the stability of the nitrenium ions of the two generated molecules and of nine further amines.

\paragraph{Setup.}
For an amine $\mathrm{ArNH}_2$ with the nitrenium ion $\mathrm{ArNH}^{+}$, we calculate the relative stability
\[
\Delta\Delta E = \bigl[E(\mathrm{ArNH}^{+}) - E(\mathrm{ArNH}_2)\bigr] - \bigl[E(\mathrm{PhNH}^{+}) - E(\mathrm{PhNH}_2)\bigr],
\]
where $\mathrm{PhNH}_2$ is aniline and $\mathrm{PhNH}^{+}$ is its nitrenium ion. A negative value means that the ion is more stable than
that of aniline relative to its amine, which is the direction that these studies associate with mutagenic amines.
\citet{furukawa2022nitrenium} and \citet{muto2024nitrenium} calculate this quantity with the semiempirical AM1 method, and we use density
functional theory instead. We optimize the geometries with the functional $\omega$B97X-D3 \citep{lin2013long} and the basis set def2-SVP
\citep{weigend2005def2}, and we calculate the energies at these geometries with def2-TZVP. Water is described by the SMD solvation model
\citep{marenich2009smd}, and the calculations run in ORCA \citep{neese2025orca} from starting geometries of GFN2-xTB
\citep{bannwarth2019gfn2}. For each ion we take the closed-shell singlet state, which lies $18$ to $42$\,kcal/mol below the triplet state that
we found. The energies are electronic energies in solution without thermal
corrections. A frequency calculation at the optimization level confirms that all $40$ optimized structures are minima. For five of them it first found one imaginary frequency, a rotation of the acetyl methyl group in the two acetamido amines and an out-of-plane motion of the hydrogen at the nitrogen in three triplet ions. We displaced these structures along that mode and optimized them again. Table~\ref{tab:nitrenium-energies} lists the absolute energies of all structures.
\ifdefined\submissionbuild The optimized geometries are provided as XYZ files in the archive \texttt{nitrenium-geometries.zip} of the Supporting Information.\fi
Besides the two generated molecules, we calculate aniline as the reference, three amines with a single substituent
(4-aminoacetanilide, o-toluidine and 4-chloroaniline), and three known potent mutagens as a scale for the effect (2-aminofluorene,
4-aminobiphenyl and 2-naphthylamine). For comparison, we add two further amines that the model generates for the non-mutagenic class
(3-aminobenzonitrile and 2-chloro-3-methylaniline).

\paragraph{Results.}
The nitrenium ion of the molecule generated as mutagenic is stabilized by $21.0$\,kcal/mol relative to that of aniline, whereas the ion of
2-chloroaniline, the molecule generated as non-mutagenic, is destabilized by $3.4$\,kcal/mol (Figure~\ref{fig:nitrenium}). The difference of
$24.4$\,kcal/mol agrees with the reading of Experiment~4, and the ion of the molecule generated as mutagenic is more stabilized than those of
the three known mutagens ($13.8$, $7.6$ and $7.2$\,kcal/mol). We attribute most of the stabilization to the acetamido group, since
4-aminoacetanilide alone reaches $-17.9$\,kcal/mol, and the rest to the methyl group next to the amine, which adds $3.1$\,kcal/mol. A chlorine atom destabilizes the ion slightly, next
to the amine ($+3.4$\,kcal/mol) and across the ring ($+0.7$\,kcal/mol), and its effect is small compared with that of the acetamido group.
The model also generates other non-mutagenic amines whose ion is less stable than that of aniline. Examples are 2-chloro-3-methylaniline
($+1.9$\,kcal/mol), with the chlorine next to the amine, and 3-aminobenzonitrile ($+10.1$\,kcal/mol), with an electron-withdrawing nitrile group.
We selected both for this property, so they show that such molecules exist and say nothing about how frequent they are.
The probability that \megan{} predicts follows the energy for most of these amines. It is at least $0.98$ for five of the six amines with a
stabilized ion and below $0.02$ for the four with a less stable ion. The exceptions are 4-aminoacetanilide ($0.33$) and aniline ($0.85$), and the
prediction disagrees with the label of the dataset for these two and for 4-chloroaniline.

\begin{figure}[t]
\centering
\includegraphics{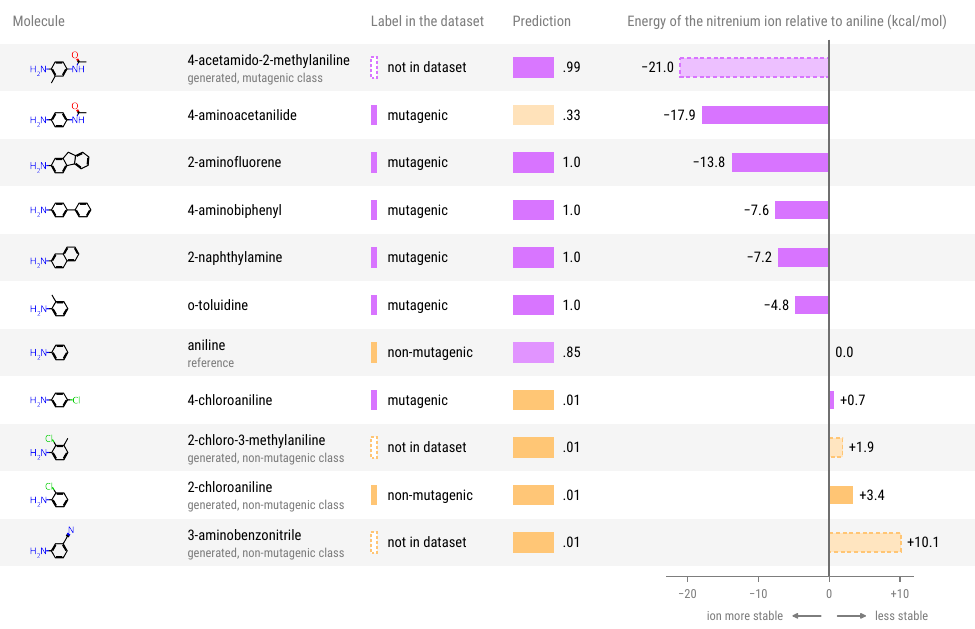}
\caption{\textbf{Stability of the nitrenium ions of eleven aromatic amines.} Each bar is the energy of the nitrenium ion relative to its amine,
minus the same difference for aniline (kcal/mol, density functional theory in water). The first color is the label of the
amine in the Ames dataset, mutagenic (purple) or non-mutagenic (orange), and the second is the class that \megan{} predicts, richer the more
certain the prediction, with the probability of mutagenicity next to it. Dashed markers and bars mark the molecules that are not in the dataset,
in the color of the class for which they were generated (the molecule generated as mutagenic in Figure~\ref{fig:e5} and two further molecules
generated as non-mutagenic). The molecule generated as non-mutagenic in Figure~\ref{fig:e5} is a compound of the dataset.}
\label{fig:nitrenium}
\end{figure}

\begin{table}[tp]
\caption{\textbf{Absolute energies of the structures of the nitrenium-ion calculations.} Electronic energies in water (SMD) in hartree, at the optimization level ($\omega$B97X-D3/def2-SVP) and at the level of the reported values ($\omega$B97X-D3/def2-TZVP single point), and the number of imaginary frequencies of a frequency calculation at the optimization level. Where two orientations of the remaining amine hydrogen are not equivalent, both singlet ions were optimized, and the lower one (marked $^{*}$) enters $\Delta\Delta E$. The amines are ordered as in Figure~\ref{fig:nitrenium}.}
\label{tab:nitrenium-energies}
\centering\footnotesize
\setlength{\tabcolsep}{4pt}
\begin{tabular*}{\linewidth}{@{\extracolsep{\fill}}llcrrc@{}}
\toprule
Amine & Structure & Charge, mult. & $E$, def2-SVP [$E_\mathrm{h}$] & $E$, def2-TZVP [$E_\mathrm{h}$] & Imag.\ freq. \\
\midrule
4-acetamido-2-methylaniline & amine & 0, 1 & $-534.432755$ & $-535.012336$ & 0 \\
 & ion, singlet$^{*}$ & +1, 1 & $-533.603771$ & $-534.183262$ & 0 \\
 & ion, singlet & +1, 1 & $-533.602976$ & $-534.183182$ & 0 \\
 & ion, triplet & +1, 3 & $-533.536803$ & $-534.116120$ & 0 \\
\addlinespace[2pt]
4-aminoacetanilide & amine & 0, 1 & $-495.151993$ & $-495.690889$ & 0 \\
 & ion, singlet & +1, 1 & $-494.317312$ & $-494.856880$ & 0 \\
 & ion, triplet & +1, 3 & $-494.269762$ & $-494.807488$ & 0 \\
\addlinespace[2pt]
2-aminofluorene & amine & 0, 1 & $-556.243615$ & $-556.825497$ & 0 \\
 & ion, singlet & +1, 1 & $-555.399376$ & $-555.984815$ & 0 \\
 & ion, singlet$^{*}$ & +1, 1 & $-555.399444$ & $-555.984866$ & 0 \\
 & ion, triplet & +1, 3 & $-555.347676$ & $-555.933130$ & 0 \\
\addlinespace[2pt]
4-aminobiphenyl & amine & 0, 1 & $-518.158870$ & $-518.704393$ & 0 \\
 & ion, singlet & +1, 1 & $-517.305664$ & $-517.853882$ & 0 \\
 & ion, triplet & +1, 3 & $-517.256546$ & $-517.804950$ & 0 \\
\addlinespace[2pt]
2-naphthylamine & amine & 0, 1 & $-440.819982$ & $-441.284516$ & 0 \\
 & ion, singlet$^{*}$ & +1, 1 & $-439.966436$ & $-440.433362$ & 0 \\
 & ion, singlet & +1, 1 & $-439.965227$ & $-440.432423$ & 0 \\
 & ion, triplet & +1, 3 & $-439.937474$ & $-440.404393$ & 0 \\
\addlinespace[2pt]
o-toluidine & amine & 0, 1 & $-326.607047$ & $-326.955111$ & 0 \\
 & ion, singlet$^{*}$ & +1, 1 & $-325.751145$ & $-326.100106$ & 0 \\
 & ion, singlet & +1, 1 & $-325.750036$ & $-326.099683$ & 0 \\
 & ion, triplet & +1, 3 & $-325.712775$ & $-326.060032$ & 0 \\
\addlinespace[2pt]
aniline & amine & 0, 1 & $-287.326584$ & $-287.634027$ & 0 \\
 & ion, singlet & +1, 1 & $-286.462458$ & $-286.771440$ & 0 \\
 & ion, triplet & +1, 3 & $-286.428171$ & $-286.734689$ & 0 \\
\addlinespace[2pt]
4-chloroaniline & amine & 0, 1 & $-746.782449$ & $-747.248003$ & 0 \\
 & ion, singlet & +1, 1 & $-745.916963$ & $-746.384365$ & 0 \\
 & ion, triplet & +1, 3 & $-745.881520$ & $-746.346531$ & 0 \\
\addlinespace[2pt]
2-chloro-3-methylaniline & amine & 0, 1 & $-786.064187$ & $-786.569178$ & 0 \\
 & ion, singlet & +1, 1 & $-785.195789$ & $-785.702352$ & 0 \\
 & ion, singlet$^{*}$ & +1, 1 & $-785.196782$ & $-785.703572$ & 0 \\
 & ion, triplet & +1, 3 & $-785.163486$ & $-785.670935$ & 0 \\
\addlinespace[2pt]
2-chloroaniline & amine & 0, 1 & $-746.782811$ & $-747.247689$ & 0 \\
 & ion, singlet & +1, 1 & $-745.912010$ & $-746.378231$ & 0 \\
 & ion, singlet$^{*}$ & +1, 1 & $-745.913194$ & $-746.379692$ & 0 \\
 & ion, triplet & +1, 3 & $-745.871383$ & $-746.338485$ & 0 \\
\addlinespace[2pt]
3-aminobenzonitrile & amine & 0, 1 & $-379.473219$ & $-379.885832$ & 0 \\
 & ion, singlet & +1, 1 & $-378.593431$ & $-379.007051$ & 0 \\
 & ion, singlet$^{*}$ & +1, 1 & $-378.593535$ & $-379.007107$ & 0 \\
 & ion, triplet & +1, 3 & $-378.561563$ & $-378.972024$ & 0 \\
\bottomrule
\end{tabular*}
\end{table}

The five amines of the dataset with a stabilized ion are all labeled mutagenic. The two amines labeled non-mutagenic, aniline and
2-chloroaniline, have no stabilized ion, and 4-chloroaniline, which is labeled mutagenic, lies between them, only $0.7$\,kcal/mol above aniline.
However, eleven amines illustrate the quantity at best, and the stability of the ion is one step of the activation among several. We therefore
read the calculation as support for the chemical reading of the example. It does not test the prediction of \megan{}, and the three generated
molecules that are not in the dataset have no experimental label. The values also depend on the functional and the solvent model, which we did not vary.
\section{Datasets}
\label{app:data}

This section describes the datasets of the experiments.

\subsection{Synthetic Colored Graphs}
\label{app:data:rbmotifs}

The synthetic dataset follows the RbMotifs construction of \citet{teufel2023megan}. It consists of $99\,995$ random
graphs of $20$ to $50$ nodes, each node carrying one of seven colors. Every graph contains between zero and eight
motifs of four types, and all but $27$ graphs contain at most six. A star is a hub with three leaves that are not
adjacent to each other, and a ring is a triangle of one node and two adjacent nodes of the same color. Each motif
type has a fixed contribution to the target value: a blue star $-2$, a blue ring $-1$, a red ring $+1$ and a red
star $+2$.

The sum of these contributions is an integer between $-3$ and $+3$ that can be recomputed exactly from the structure
of a graph. The dataset is balanced, with $14\,285$ graphs for each of the seven values, and the label of a graph is
this sum plus uniform noise of at most $0.5$. The same value can be reached by different combinations of motifs, for
example $+2$ by a single red star or by two red rings.
Figure~\ref{fig:data-synthetic} shows the motifs, an example graph and the distribution of the label.

\begin{figure}[t]
\centering
\includegraphics{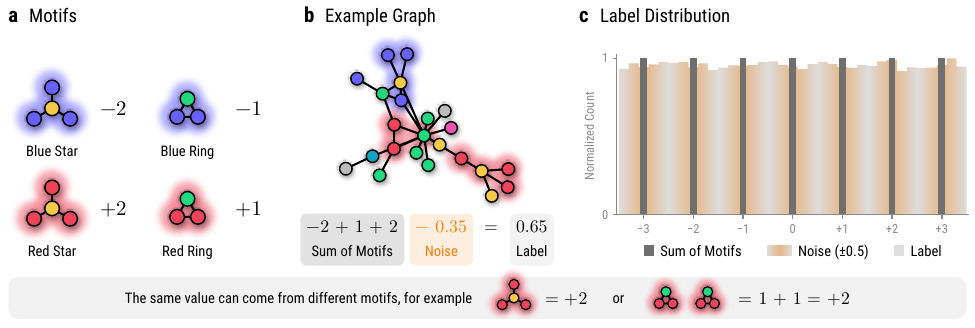}
\caption{\textbf{The synthetic colored graphs.} (a) The four motif types with their contribution to the target value.
A star is a yellow hub with three leaves that are not adjacent to each other, and a ring is a triangle of a green
node and two adjacent nodes of the same color, blue for a negative and red for a positive contribution. (b) A graph
of the dataset with $23$ nodes, whose motifs are marked by the colored glow. The three motifs sum to $+1$, and with
a noise of $-0.35$ the label of the graph is $0.65$. The nodes have one of seven colors (gray, red, green, blue,
yellow, cyan and magenta). (c) Distribution of the sum of the motifs (dark) and of the label (light) over all
$99\,995$ graphs, each normalized to its maximum. The dataset is balanced over the seven values of the sum, and the
noise moves the label of a graph by at most $0.5$ to either side of its sum (orange).}
\label{fig:data-synthetic}
\end{figure}

\subsection{ZINC-250k and the Labeled Subset}
\label{app:data:zinc}

ZINC \citep{sterling2015zinc15} is a public database of commercially available, drug-like molecules. The ZINC-250k
selection contains about $250$k of them, with at most $38$ heavy atoms. Molecules are written as graphs whose nodes
are atoms and whose edges are bonds, over the nine elements that occur in the data.

A subset of $65\,434$ molecules carries a label, the Wildman--Crippen estimate of the octanol--water partition
coefficient $\log P$ \citep{wildman1999clogp}. It is a deterministic function of the molecular graph, which we
compute with RDKit. The values range from $-3.7$ to $7.8$ with a mean of $2.8$.

\subsection{ZINC and ChEMBL Corpus}
\label{app:data:corpus}

The corpus contains $1\,027\,984$ molecular graphs. It combines the ZINC molecules above with the molecules of
ChEMBL \citep{zdrazil2024chembl} (release~37), a public database of bioactive compounds, that have at most
$24$ heavy atoms. Protonation is neutralized. In contrast to the molecules above, charged atoms are kept, with the
three additional atom classes N$^{+}$, O$^{-}$ and N$^{-}$, so that groups such as nitro, azide and N-oxide groups
can be written. We assembled the corpus without reference to the Ames data below.

\subsection{Ames Mutagenicity}
\label{app:data:ames}

The Ames benchmark of \citet{hansen2009ames} collects molecules that have been tested for mutagenicity in the Ames
assay, which exposes the bacterium \textit{Salmonella typhimurium} to a compound and records whether it causes
mutations. Each molecule carries a binary label that marks the mutagenic class. Our copy lists $6506$ compounds.
After we remove the compounds with more than $38$ heavy atoms or with elements outside the nine of ZINC and merge
duplicates, none of which carry conflicting labels, $6301$ distinct graphs remain. Of these, $3424$
($54.3\%$) are mutagenic, and the median graph has $16$ heavy atoms.
\section{Experiment Details}
\label{app:protocol}

This section describes how we measure the results of \Secref{sec:experiments}. The first part defines the matches,
the target cells, the arms and the settings that the three computational rungs share, and the second part
describes the real-world rung on mutagenicity, whose results cannot be graded by an independent oracle. The third part
states how we find the concept prototypes of \appref{app:prototypes}.

\subsection{Matches, Cells and Settings}
\label{app:protocol:metrics}

\paragraph{Matches.}
A graph meets the property condition if $|f(G) - y^*| \le \tau$, where $f$ is the deterministic function that
defines the property and $y^*$ the requested value. The tolerance is $\tau = 1$ for the synthetic colored graphs, whose
property is an integer that we recompute from the structure of a generated graph, and $\tau = 0.5$ for the Crippen
$\log P$ of the molecular graphs. A graph meets the concept condition if its membership satisfies $\phic < 0.5$ in
the \megan{} model that defines the concept (\appref{app:megan}), and a joint match requires both conditions.
On the molecular rungs, a match also requires a valid graph of a single fragment, and every rate counts every graph that was generated, so that an invalid graph is a
miss. We report each rate next to the rate of the same condition among unconditional samples, since the tolerance band
of a property or the core of a concept can be large enough to be hit by chance, and we report the
validity, the share of distinct graphs and the cost of a distinct joint match, which is the median number of
samples that have to be drawn from an arm to obtain one distinct graph that meets both conditions, over the supported
cells.

\paragraph{Target cells.}
A target cell is a concept together with a requested value of the property, and every cell is sampled with
$1024$ graphs. The synthetic rung uses six concepts and the five values $\{-3, -1, 0, +1, +3\}$, and the molecular
rung uses six concepts and the five values $\log P \in \{0.5, 1.5, 2.5, 3.5, 4.5\}$, which gives $30$ cells each. We
classify a cell by how many of the core members of its concept, the graphs with $\phic$ near $0$, carry the
requested value. A cell is native if at least $15\%$ of the core members carry it and unsupported if none do on the synthetic
rung, where the value is exact, or fewer than ten do on the molecular rung, where the Crippen value of a member lies
within $0.5$ of the request. A cell in between is off-native. In an off-native cell, the data contain
few examples of the combination, so that a method that reaches it does more than reproduce a training distribution.
The cost of a distinct joint match is taken over the native and off-native cells, $27$ of the $30$ on both rungs, and the
unsupported cells enter the other columns.

\paragraph{Arms and settings.}
We compare the base without conditioning ($1024$ graphs), the property adapter alone and the concept adapter alone
(per level and per concept respectively), both adapters, Feynman--Kac steering of the base alone and both adapters
with steering, the last three on $1024$ graphs per cell. On the molecular rungs, the sampler takes $500$ steps with
a stochasticity of $\eta = 500$, and on the synthetic rung the stochasticity is $\eta = 0$, so that the absolute values of the two are not
comparable. Steering uses $k = 16$ particles and the energy of
\Secref{sec:method:fk}, and on the molecular rungs $\beta = 16$ with resampling over the last $30\%$ of the
trajectory. We fix the guidance weights by rules that do not look at the steered arms. For a single adapter, we take
the weight with the highest match when invalid graphs count as misses, and for the composition of both adapters, we
take the pair $(w_c, w_p)$ with the highest mean joint match over the native and off-native cells among a small
grid of pairs, before any steering run. This grid consists of the pairs $(1, 2)$, $(1, 1)$ and $(0.5, 1)$ for
Experiment~2 and of $(1, 2)$, $(1, 1)$ and $(2, 2)$ for Experiment~3, and the rule selects $(1, 1)$ in both cases.
\Secref{app:fk_adapter:sweep} relates this choice to a larger grid of pairs for Experiment~2.

\subsection{Real-World Mutagenicity}
\label{app:protocol:mutagenicity}

\paragraph{Data, grader and concepts.}
The real-world rung uses the Ames benchmark \citep{hansen2009ames} (\appref{app:data:ames}). Since the base of the
other experiments cannot write charged groups such as the nitro group, we train a second frozen base on the ZINC and
ChEMBL corpus (\appref{app:data:corpus}). The grader is a \megan{} model that we retrain as a classifier on the Ames
graphs that this base can write. Table~\ref{tab:megan-quality} in \appref{app:perconcept} reports its quality,
including the fidelity check that its non-mutagenic channel fails. Unlike the Crippen $\log P$, the mutagenicity of a
generated molecule cannot be graded by an oracle. The same model defines the concepts and, in the steered arms, supplies
the energy. A match therefore measures how well an arm steers the prediction of this model and not whether the molecule
is mutagenic. We read every rate of this rung as a comparison between arms that one and the same model grades. We use
seven concepts, each named by the structural group that dominates its core members, namely N-nitrosamines, epoxides,
azides, nitro-heteroarenes, azo compounds, aromatic amines and sulfonyl compounds. Figure~\ref{fig:e5} shows example
molecules that we select from the steered samples of these concepts.

\paragraph{Cells, arms and weights.}
We request each concept for both classes, which yields $14$ cells, and the property is the probability of
mutagenicity that \megan{} predicts. A property match requires a valid molecule whose predicted probability lies
within $0.5$ of the level, that is, whose predicted class is the requested one. We classify the cells as in
\Secref{app:protocol:metrics}. Of the $14$ cells, $8$ are native, $1$ is off-native (the aromatic amines at the
non-mutagenic class) and $5$ are unsupported.

The arms are those of the other rungs, with the two classes as the levels $0$ and $1$, all at $\eta = 500$. An arm that
does not receive the class, the base and the concept adapter, is graded at both classes, and every valid graph meets
exactly one of them. Its property match is therefore half its validity and its joint match half its concept match,
which is the rate at which it meets both conditions when it is asked for both and flatters its cost accordingly. We use
the class adapter at $w = 1.5$ and the concept adapter at $w = 1$, each with the highest match when invalid graphs
count as misses, and the composition at $(w_c, w_p) = (1, 1)$, which the rule of \Secref{app:protocol:metrics} selects
before any steering run.

\subsection{Concept Prototype Search}
\label{app:protocol:prototypes}

\appref{app:prototypes} asks how small a graph can be and still belong to a concept, and it answers by a sweep
over the number of nodes. For a concept and a size, we draw $128$ graphs of exactly that size from the concept adapter
alone at $w = 1$, with the frozen base, the \megan{} model and the sampler settings of Experiment~4 ($500$ steps,
$\eta = 500$). A graph is a hit if it is valid, connected, plausible and has $\phic < 0.5$, and the hit rate is the share of hits. We sweep from the median size of the
core members of the concept down to $3$ nodes and take the plateau as the median hit rate over the largest third of
the sizes. The elbow is the smallest size at which the hit rate still reaches half of the plateau, and the prototype is
the most frequent distinct hit at the elbow. Two checks do not rely on the adapter or on the grader. The frozen base
alone draws $256$ graphs per size, and we count the graphs that contain the defining substructure of the concept.
\appref{app:prototypes} gives the equation of the elbow and the results for the seven named concepts.
\section{Detailed Results}
\label{app:perconcept}

\renewcommand{\topfraction}{0.75}\renewcommand{\textfraction}{0.25}\renewcommand{\floatpagefraction}{0.65}

This section reports the results of Experiments~1 to~4 in more detail than the main text. For each experiment, we extend
the tables of the main text by the standard deviation over the cells on which their means are taken, and we resolve the
aggregates by requested level, by concept and by target cell, in a table or in a heatmap. \Secref{app:perconcept:single}
covers the individual adapters on the synthetic graphs (Experiment~1), \Secref{app:perconcept:synthetic} the joint
conditioning on these graphs (Experiment~2) and \Secref{app:perconcept:molecular} the molecular graphs (Experiment~3).
\Secref{app:perconcept:realworld} reports the arms of the real-world experiment (Experiment~4), which the main text
discusses qualitatively, and \Secref{app:perconcept:megan} gives the quality of the \megan{} models that define the concepts
and grade the matches.

Every number of this section stems from one sampling run with one trained adapter of each kind. The standard deviations
therefore describe how much the target cells differ from each other, and not how much the results vary between repeated
runs, and we set them as small numbers behind the means. A first run of the joint experiments with $128$ graphs per cell gives an
indication of that variation. Its means of the joint match differ from those reported here by at most $0.03$, whereas single cells
of the arms with steering differ by up to $0.43$, against at most $0.08$ for the other arms. We believe that the resampling of the
particles makes the graphs of a steered run depend on each other. Single cells of these arms are therefore less certain than
the binomial error of their sample suggests. Since the cells differ strongly, we additionally compare the arms
cell by cell, by counting the cells in which one arm exceeds another. We count only supported cells, that is, cells for which
the training data hold enough graphs with the requested combination, as defined in \appref{app:protocol}. As in the
main text, the best mean of each column is highlighted in bold. We name concepts by their identifiers, where the prefix ch0 or
ch1 gives the explanation channel of \megan{} to which a concept belongs.

\subsection{Individual Adapters on the Synthetic Graphs}
\label{app:perconcept:single}

\paragraph{Property adapter.}
Table~\ref{tab:g-e1} extends Table~\ref{tab:e1} by the standard deviation over the seven requested levels and by the
divergence of the generated graphs from the pool. The cross-attention adapter reaches its target about equally well at
every level, since its error varies by only $0.03$ over the levels at both weights, whereas the error varies by $0.61$ for
the base and by $0.21$ for the FiLM-only conditioning. Table~\ref{tab:e1-levels} shows where the latter variation comes from.
The error of the FiLM-only conditioning is lowest at the extreme levels ($0.81$ at $-3$ and $0.78$ at $+3$) and highest in
the middle of the range ($1.27$ at $0$), which is the reverse of the base, whose error is highest at the extremes. We believe
that this reflects the many-to-one construction of the property, in which the middle values can be composed from more
combinations of motifs than the extreme ones. The cross-attention adapter shows no such dependence on the level, and steering
lowers its error to $0.04$ or less at every level. However, we show the FiLM-only conditioning at $w = 2$, the better of the
two weights that we evaluated ($1.32$ at $w = 1$), so that we cannot say whether other weights remove its dependence on the
level. Steering also raises the divergence from the pool from $0.007$ to $0.107$, almost entirely through the sizes
($0.105$), and the mean number of nodes falls from $32.5$ to $30.2$, whereas the adapters alone do not increase the
divergence ($0.005$).

\begin{table}[tp]
\centering
\footnotesize
\setlength{\tabcolsep}{3.2pt}
\caption{\textbf{Property adapter on the synthetic graphs, with the spread over the levels and the divergence from the
pool.} Table~\ref{tab:e1} with one further row for the full adapter at $w = 1$, and with the standard deviation over the seven requested levels as small numbers ($1024$ graphs per
level, and one sample of $7168$ graphs for the base), and the Kullback--Leibler divergence of the colors, degrees and sizes
of the generated graphs from those of $5000$ pool graphs, in units of $10^{-3}$ and over all levels together, with the mean
number of nodes. The conditional arms hold the same number of graphs at every level, whereas the pool does not. MAE against
the exact motif sum, and the best mean per column in bold.}
\label{tab:g-e1}
\begin{tabular*}{\linewidth}{@{\extracolsep{\fill}} l ccc cccc c}
\toprule
 & & & & \multicolumn{4}{c}{KL divergence ($10^{-3}$)} & \\
\cmidrule(lr){5-8}
Model & MAE $\downarrow$ & Unique $\uparrow$ & Valid $\uparrow$ & Colors & Degrees & Sizes & Total $\downarrow$ & Nodes \\
\midrule
Base (no conditioning) & 2.271{\scriptsize$\pm$.612} & \textbf{1.000}{\scriptsize$\pm$.000} & 0.992{\scriptsize$\pm$.001} & 0.11 & 0.55 & 6.11 & 6.78 & 32.5 \\
Adapter, FiLM only & 1.076{\scriptsize$\pm$.208} & \textbf{1.000}{\scriptsize$\pm$.000} & 0.994{\scriptsize$\pm$.004} & 0.05 & 0.55 & 4.55 & 5.15 & 32.5 \\
Adapter, full ($w{=}1$) & 0.455{\scriptsize$\pm$.030} & \textbf{1.000}{\scriptsize$\pm$.000} & 0.994{\scriptsize$\pm$.002} & 0.10 & 0.40 & 4.26 & 4.76 & 32.5 \\
Adapter, full ($w{=}2$) & 0.148{\scriptsize$\pm$.026} & \textbf{1.000}{\scriptsize$\pm$.000} & 0.994{\scriptsize$\pm$.003} & 0.10 & 0.36 & 4.18 & \textbf{4.64} & 32.3 \\
Adapter $+$ FK & \textbf{0.026}{\scriptsize$\pm$.008} & 0.812{\scriptsize$\pm$.017} & \textbf{0.999}{\scriptsize$\pm$.001} & 0.68 & 1.34 & 104.71 & 106.73 & 30.2 \\
\bottomrule
\end{tabular*}

\end{table}

\paragraph{Concept adapter.}
Table~\ref{tab:e1-concepts} resolves the concept adapter at $w = 1$ by concept, with $1024$ graphs per concept. The adapter
reaches every concept far more often than the base does, with matches between $0.35$ and $0.88$ (mean $0.72$) and a lift of
at least $0.34$ for all $19$ concepts, and the standard error of a single match is at most $0.016$. The three lowest matches
are those of ch1 cl17 ($0.35$), ch1 cl16 ($0.44$) and ch0 cl0 ($0.47$). However, a match is the membership that the
\megan{} model assigns, and it does not check the defining motifs of the concept.

The last column of Table~\ref{tab:e1-concepts} gives the other side of the request. The property of the generated graphs
deviates from that of the real members of the concept by $1.2$ on average, which is $20\%$ of the range of the property, and
the deviation points in the direction of the channel for $18$ of the $19$ concepts, the exception being ch0 cl0 ($+0.64$).
Table~\ref{tab:g-e1-effects} adds the spread to Table~\ref{tab:e1-effects}. The concept adapter shifts the property by
$20.3\%$ of its range, with a standard deviation of $6.9$ percentage points over the concepts, whereas the property adapter
shifts the concept rates by $1.2\%$, which is about $17$ times less. Even the smallest shift of a concept ($0.50$, or $8.4\%$
of the range) is more than seven times that of the property adapter, which indicates that the asymmetry of the main text does
not rest on a few concepts.

\begin{table}[tp]
\centering
\scriptsize
\setlength{\tabcolsep}{2.6pt}
\captionsetup{font=footnotesize}
\begin{minipage}[t]{0.47\textwidth}
\centering
\captionof{table}{\textbf{Concept adapter ($w = 1$) per concept.} Each concept holds $1024$ graphs, and a match requires
$\phic < 0.5$. Unc is the same share among $1024$ unconditional samples, lift the difference, and the shift the mean property
of the generated graphs minus that of the real core members of the concept (the property has a range of $6$). The last rows
give the mean and the standard deviation over the concepts, with absolute values for the shift.}
\label{tab:e1-concepts}
\begin{tabular*}{\linewidth}{@{\extracolsep{\fill}} l ccc c}
\toprule
Concept & Match & Unc & Lift & Property shift \\
\midrule
ch0 cl0 & 0.467 & 0.005 & +0.462 & +0.64 \\
ch0 cl1 & 0.720 & 0.027 & +0.692 & $-$1.57 \\
ch0 cl2 & 0.770 & 0.143 & +0.627 & $-$1.51 \\
ch0 cl3 & 0.870 & 0.147 & +0.723 & $-$0.82 \\
ch0 cl4 & 0.790 & 0.075 & +0.715 & $-$0.54 \\
ch0 cl5 & 0.777 & 0.011 & +0.767 & $-$1.16 \\
ch0 cl6 & 0.803 & 0.017 & +0.786 & $-$1.77 \\
ch0 cl7 & 0.737 & 0.157 & +0.580 & $-$1.02 \\
ch0 cl8 & 0.789 & 0.079 & +0.710 & $-$1.51 \\
ch1 cl9 & 0.741 & 0.015 & +0.727 & +1.36 \\
ch1 cl10 & 0.759 & 0.016 & +0.743 & +0.78 \\
ch1 cl11 & 0.755 & 0.074 & +0.681 & +0.50 \\
ch1 cl12 & 0.829 & 0.199 & +0.630 & +0.83 \\
ch1 cl13 & 0.763 & 0.018 & +0.745 & +1.57 \\
ch1 cl14 & 0.883 & 0.081 & +0.802 & +1.46 \\
ch1 cl15 & 0.789 & 0.027 & +0.762 & +1.30 \\
ch1 cl16 & 0.441 & 0.037 & +0.404 & +1.49 \\
ch1 cl17 & 0.353 & 0.009 & +0.344 & +1.64 \\
ch1 cl18 & 0.698 & 0.080 & +0.618 & +1.61 \\
\midrule
mean & 0.723 & 0.064 & $\mathbf{+0.659}$ & 1.22 \\
std & 0.143 & 0.059 & 0.130 & 0.41 \\
\bottomrule
\end{tabular*}

\end{minipage}\hfill
\begin{minipage}[t]{0.49\textwidth}
\centering
\captionof{table}{\textbf{What each adapter changes, with the spread.} Table~\ref{tab:e1-effects} with the standard
deviation over the seven requested levels (property adapter) and over the $19$ concepts (concept adapter); in the base
row over the levels for the property match and over the concepts for the concept match.}
\label{tab:g-e1-effects}
\begin{tabular*}{\linewidth}{@{\extracolsep{\fill}} l cc cc}
\toprule
 & \multicolumn{2}{c}{Match $\uparrow$} & \multicolumn{2}{c}{Shift (\%) $\downarrow$} \\
\cmidrule(lr){2-3}\cmidrule(lr){4-5}
 & property & concept & property & concept \\
\midrule
Base & 0.39{\scriptsize$\pm$.12} & 0.06{\scriptsize$\pm$.06} & --- & --- \\
Property adapter & \requested{0.87{\scriptsize$\pm$.01}} & 0.06{\scriptsize$\pm$.01} & --- & 1.2{\scriptsize$\pm$.3} \\
Concept adapter & 0.39{\scriptsize$\pm$.05} & \requested{0.72{\scriptsize$\pm$.14}} & 20.3{\scriptsize$\pm$6.9} & --- \\
\bottomrule
\end{tabular*}

\vspace{1.4em}
\captionof{table}{\textbf{Error of the property per requested level.} Mean absolute error against the exact motif sum,
with $1024$ graphs per level and one sample of $7168$ graphs for the base.}
\label{tab:e1-levels}
\begin{tabular*}{\linewidth}{@{\extracolsep{\fill}} r ccccc}
\toprule
$y^*$ & Base & FiLM & $w{=}1$ & $w{=}2$ & FK \\
\midrule
$-3$ & 3.068 & 0.811 & 0.460 & 0.153 & \textbf{0.026} \\
$-2$ & 2.293 & 1.062 & 0.442 & 0.177 & \textbf{0.027} \\
$-1$ & 1.779 & 1.242 & 0.444 & 0.176 & \textbf{0.029} \\
$0$ & 1.587 & 1.271 & 0.454 & 0.160 & \textbf{0.032} \\
$+1$ & 1.786 & 1.269 & 0.498 & 0.137 & \textbf{0.033} \\
$+2$ & 2.299 & 1.095 & 0.479 & 0.117 & \textbf{0.027} \\
$+3$ & 3.088 & 0.783 & 0.404 & 0.114 & \textbf{0.009} \\
\bottomrule
\end{tabular*}

\end{minipage}
\end{table}

\subsection{Joint Conditioning on the Synthetic Graphs}
\label{app:perconcept:synthetic}

Table~\ref{tab:g-e2} extends Table~\ref{tab:e2} by the standard deviation over the $30$ target cells and by the number of
supported cells without a distinct joint match. The deviations are large compared to the differences between neighboring
arms, for example $0.38$ for the joint match of the best arm ($0.54$), since the cells differ strongly in how well the
training data support them. We therefore also compare the arms cell by cell. The composed adapters with steering reach a
higher joint match than the composed adapters alone in all $27$ supported cells, and than steering alone in $18$ of
them. Steering alone is ahead in eight cells, which are the levels $-1$, $0$ and $+1$ of ch0 cl3, the level $-3$ of ch0 cl7 and
the levels $-3$, $-1$, $0$ and $+1$ of ch1 cl12, so that the order of these two arms does not hold in every cell. With $1024$
graphs per cell, the composed adapters, steering alone and both together find a distinct joint match in every supported
cell, whereas the base misses one cell, the concept adapter two and the property adapter three.

\begin{table}[tp]
\centering
\footnotesize
\setlength{\tabcolsep}{1.5pt}
\caption{\textbf{Joint conditioning on the synthetic graphs, with the spread over the cells.}
Table~\ref{tab:e2} with the standard deviation over the $30$ target cells as small numbers, and over the
cells of the respective class for the native and the off-native columns. The cost is the median number of
samples per distinct joint match over the $27$ supported cells, and a dash marks an arm without any in more
than half of them. The last column counts the supported cells without a distinct joint match.}
\label{tab:g-e2}
\begin{tabular*}{\linewidth}{@{\extracolsep{\fill}} l ccc cc cc r c}
\toprule
 & \multicolumn{3}{c}{Match $\uparrow$} & \multicolumn{2}{c}{Joint by cell $\uparrow$} & & & & \\
\cmidrule(lr){2-4}\cmidrule(lr){5-6}
Model & Prop. & Concept & Joint & Native & Off-nat. & Unique $\uparrow$ & Valid $\uparrow$ & Cost $\downarrow$ & No hit $\downarrow$ \\
\midrule
Base (no conditioning) & 0.392{\scriptsize$\pm$.137} & 0.119{\scriptsize$\pm$.055} & 0.044{\scriptsize$\pm$.042} & 0.086{\scriptsize$\pm$.034} & 0.027{\scriptsize$\pm$.027} & \textbf{1.000}{\scriptsize$\pm$.000} & 0.995{\scriptsize$\pm$.000} & 25.6 & 1/27 \\
$+$ property adapter & \textbf{0.956}{\scriptsize$\pm$.009} & 0.119{\scriptsize$\pm$.188} & 0.114{\scriptsize$\pm$.182} & 0.298{\scriptsize$\pm$.221} & 0.026{\scriptsize$\pm$.028} & \textbf{1.000}{\scriptsize$\pm$.000} & 0.993{\scriptsize$\pm$.004} & 26.3 & 3/27 \\
$+$ concept adapter & 0.321{\scriptsize$\pm$.387} & \textbf{0.749}{\scriptsize$\pm$.140} & 0.222{\scriptsize$\pm$.309} & 0.540{\scriptsize$\pm$.292} & 0.074{\scriptsize$\pm$.172} & \textbf{1.000}{\scriptsize$\pm$.000} & 0.991{\scriptsize$\pm$.004} & 21.8 & 2/27 \\
$+$ both adapters & 0.541{\scriptsize$\pm$.313} & 0.500{\scriptsize$\pm$.252} & 0.276{\scriptsize$\pm$.328} & 0.646{\scriptsize$\pm$.274} & 0.107{\scriptsize$\pm$.152} & \textbf{1.000}{\scriptsize$\pm$.000} & 0.993{\scriptsize$\pm$.003} & 7.2 & \textbf{0}/27 \\
$+$ FK ($k{=}16$) & 0.873{\scriptsize$\pm$.066} & 0.481{\scriptsize$\pm$.287} & 0.405{\scriptsize$\pm$.287} & 0.690{\scriptsize$\pm$.108} & 0.309{\scriptsize$\pm$.227} & 0.760{\scriptsize$\pm$.030} & \textbf{0.999}{\scriptsize$\pm$.001} & 2.7 & \textbf{0}/27 \\
$+$ both adapters $+$ FK & 0.732{\scriptsize$\pm$.221} & 0.741{\scriptsize$\pm$.305} & \textbf{0.544}{\scriptsize$\pm$.375} & \textbf{0.932}{\scriptsize$\pm$.104} & \textbf{0.412}{\scriptsize$\pm$.285} & 0.817{\scriptsize$\pm$.041} & \textbf{0.999}{\scriptsize$\pm$.001} & \textbf{2.2} & \textbf{0}/27 \\
\bottomrule
\end{tabular*}

\end{table}

Figure~\ref{fig:g-heat-syn} shows the three match rates for every cell. Each adapter reaches its own condition whatever the
other one asks for, since the property adapter lies at $0.94$ to $0.96$ in every cell of the property match and the concept
adapter between $0.46$ and $0.87$ for the six concepts. The other condition, however, follows the data. For ch0 cl3, the
concept match of the property adapter falls from $0.70$ at $-3$ to $0.00$ at $+3$, and the property match of the concept
adapter falls from $0.94$ to $0.00$ over the same levels. The joint match of the composed adapters accordingly concentrates in
the native cells, and steering extends it to cells that the data support only weakly. In the cell of ch0 cl7 at $+3$, which
holds $287$ graphs of the pool ($2.7\%$ of the core members), the joint match rises from $0.15$ for the composed adapters to
$0.52$ for steering alone and $0.64$ for both together.

\begin{figure}[tp]
\centering
\includegraphics[width=\textwidth]{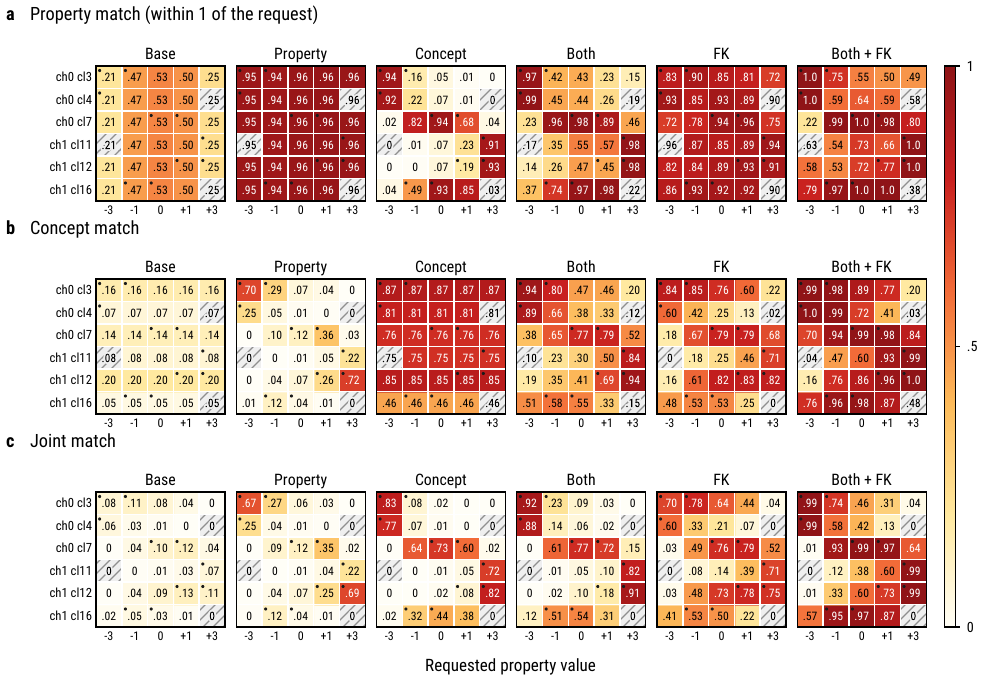}
\caption{\textbf{Joint conditioning on the synthetic graphs, per target cell.} Share of the generated
graphs that \textbf{a}~have a property within $1$ of the request, \textbf{b}~are core members of the requested
concept and \textbf{c}~satisfy both, for the six arms of Table~\ref{tab:e2} (columns), per concept (rows) and
requested value. A dot marks a native cell, and hatching a cell without any graph of the training data
with that combination. The joint arms hold $1024$ graphs per cell, whereas the base (one sample), the property adapter (one sample per
level) and the concept adapter (one sample per concept) are graded against every cell that they apply to, so that
their cells share graphs. All panels share one color scale.}
\label{fig:g-heat-syn}
\end{figure}

However, no arm exceeds a joint match of $0.04$ in the cell of ch0 cl3 at $+3$. Steering alone reaches the
requested value in $0.72$ of the graphs there, and the composed adapters with steering reach the concept
in $0.20$. The pool contains only $15$ graphs ($0.08\%$ of the core members of the concept) with this
combination. We therefore believe that the combination is too rare in the data for any of the arms to
reach it more often. Every cell holds $1024$ graphs, which gives a sampling error of up to $0.03$ for a
single rate ($95\%$ interval) of the arms without steering, so that differences of less than approximately $0.05$
between single cells of these arms should not be interpreted.

\subsection{Molecular Graphs}
\label{app:perconcept:molecular}

\paragraph{Individual adapters.}
Table~\ref{tab:e3} extends the evaluation of the individual adapters on molecular graphs by the standard deviation, which is
taken over the five requested levels for the property adapter and over the $42$ concepts for the concept adapter, and
Table~\ref{tab:g-mol-levels} lists the match of the property adapter by level. The property adapter reaches the requested
$\log P$ at every level, with a match between $0.81$ and $0.87$ at $w = 2$, whereas the match of the base never exceeds $0.31$
and is highest at the middle levels. The weight $w = 3$ lowers the validity to $0.57$ and the match to $0.47$, which is why
Experiment~3 uses $w = 2$.

\begin{table}[tp]
\centering
\scriptsize
\setlength{\tabcolsep}{2.8pt}
\captionsetup{font=footnotesize}
\begin{minipage}[t]{0.56\textwidth}
\centering
\captionof{table}{\textbf{Individual adapters on molecular graphs, with the spread.} The property adapter is evaluated at
five levels of $\log P$, where a match lies within $0.5$ of the request, and the concept adapter on $42$ concepts with $1024$
molecules each. The small numbers give the standard deviation over the levels (property adapter and base) or over the
concepts (concept adapter, and the concept match of the base). A dash marks a quantity that an arm does not target.}
\label{tab:e3}
\begin{tabular*}{\linewidth}{@{\extracolsep{\fill}} l cccc}
\toprule
Model & Valid $\uparrow$ & $\log P$ $\uparrow$ & MAE $\downarrow$ & Concept $\uparrow$ \\
\midrule
Base & \textbf{0.988}{\scriptsize$\pm$.000} & 0.192{\scriptsize$\pm$.111} & 1.50{\scriptsize$\pm$.54} & 0.007{\scriptsize$\pm$.008} \\
\addlinespace[1pt]
Property, $w{=}1$ & 0.981{\scriptsize$\pm$.007} & 0.802{\scriptsize$\pm$.027} & 0.30{\scriptsize$\pm$.02} & --- \\
Property, $w{=}2$ & 0.946{\scriptsize$\pm$.012} & \textbf{0.844}{\scriptsize$\pm$.028} & \textbf{0.25}{\scriptsize$\pm$.02} & --- \\
Property, $w{=}3$ & 0.569{\scriptsize$\pm$.072} & 0.471{\scriptsize$\pm$.048} & 0.29{\scriptsize$\pm$.02} & --- \\
\addlinespace[1pt]
Concept, $w{=}1$ & 0.979{\scriptsize$\pm$.012} & --- & --- & 0.245{\scriptsize$\pm$.151} \\
Concept, $w{=}2$ & 0.851{\scriptsize$\pm$.165} & --- & --- & \textbf{0.361}{\scriptsize$\pm$.185} \\
\bottomrule
\end{tabular*}

\end{minipage}\hfill
\begin{minipage}[t]{0.40\textwidth}
\centering
\captionof{table}{\textbf{Property adapter on molecular graphs, per requested level.} Match within $0.5$ of the requested
$\log P$ for the base and the property adapter at three weights. The last rows give the mean and the standard deviation over
the levels.}
\label{tab:g-mol-levels}
\begin{tabular*}{\linewidth}{@{\extracolsep{\fill}} r cccc}
\toprule
$\log P$ & Base & $w{=}1$ & $w{=}2$ & $w{=}3$ \\
\midrule
$0.5$ & 0.048 & 0.758 & \textbf{0.807} & 0.516 \\
$1.5$ & 0.191 & 0.795 & \textbf{0.862} & 0.531 \\
$2.5$ & 0.307 & 0.812 & \textbf{0.861} & 0.436 \\
$3.5$ & 0.292 & 0.826 & \textbf{0.867} & 0.440 \\
$4.5$ & 0.121 & 0.817 & \textbf{0.822} & 0.432 \\
\midrule
mean & 0.192 & 0.802 & 0.844 & 0.471 \\
std & 0.111 & 0.027 & 0.028 & 0.048 \\
\bottomrule
\end{tabular*}

\end{minipage}
\end{table}

\paragraph{Concept adapter.}
The concept adapter raises the concept match above that of the base for all $42$ concepts at both weights
(Table~\ref{tab:g-mol-concepts}), with matches between $0.01$ and $0.59$ at $w = 1$ and between $0.02$ and $0.69$ at $w = 2$.
The larger weight comes at the price of validity, which falls below $0.5$ for two concepts (ch0 cl14 with $0.23$ and ch0 cl13
with $0.34$), so that its standard deviation over the concepts grows from $0.01$ to $0.17$. The concepts of channel 1 are
matched more often than those of channel 0 ($0.28$ against $0.18$ at $w = 1$ and $0.41$ against $0.27$ at $w = 2$). The
smallest gain is that of ch1 cl15, whose match rises only from $0.00$ to $0.01$ at $w = 1$ and to $0.02$ at $w = 2$.

Requesting a concept also moves the property of the molecules. At $w = 1$, the $\log P$ of the generated molecules is lower than
that of the real members of the concept for $34$ of the $42$ concepts, including all $15$ concepts of channel 0. The
mean absolute shift is $0.28$, which grows to $0.53$ at $w = 2$. The largest shift occurs for ch0 cl7 ($-1.04$ at $w = 1$ and
$-2.69$ at $w = 2$). In contrast, requesting the property leaves the concept rates almost where the pool puts them, with a mean
absolute shift of $0.003$ at $w = 1$ and of $0.004$ at $w = 2$. This is the same asymmetry that we find on the synthetic graphs.

\begin{table}[tp]
\centering
\scriptsize
\setlength{\tabcolsep}{2.6pt}
\caption{\textbf{Concept adapter on molecular graphs, per concept.} Match of the requested concept for the base and the concept
adapter at $w = 1$ and $w = 2$ ($1024$ molecules per concept), and the shift $\Delta \log P$ of the mean $\log P$ of the generated
molecules relative to the real members of the concept at both weights. The last row gives the mean over the concepts, with the
absolute value for the shift.}
\label{tab:g-mol-concepts}
\begin{tabular*}{\linewidth}{@{\extracolsep{\fill}} l ccc cc @{\hspace{10pt}} l ccc cc}
\toprule
 & \multicolumn{3}{c}{Match} & \multicolumn{2}{c}{$\Delta\log P$} & & \multicolumn{3}{c}{Match} & \multicolumn{2}{c}{$\Delta\log P$} \\
\cmidrule(lr){2-4}\cmidrule(lr){5-6}\cmidrule(lr){8-10}\cmidrule(lr){11-12}
Concept & Base & $w{=}1$ & $w{=}2$ & $w{=}1$ & $w{=}2$ & Concept & Base & $w{=}1$ & $w{=}2$ & $w{=}1$ & $w{=}2$ \\
\midrule
ch0 cl0 & 0.002 & 0.046 & 0.077 & $-$0.40 & $-$0.78 & ch1 cl21 & 0.001 & 0.203 & 0.274 & +0.01 & $-$0.06 \\
ch0 cl1 & 0.012 & 0.160 & 0.411 & $-$0.42 & $-$0.74 & ch1 cl22 & 0.036 & 0.456 & 0.633 & $-$0.55 & $-$0.61 \\
ch0 cl2 & 0.009 & 0.161 & 0.470 & $-$0.32 & $-$0.72 & ch1 cl23 & 0.008 & 0.103 & 0.160 & $-$0.07 & $-$0.15 \\
ch0 cl3 & 0.009 & 0.200 & 0.438 & $-$0.46 & $-$0.76 & ch1 cl24 & 0.002 & 0.301 & 0.464 & $-$0.07 & $-$0.05 \\
ch0 cl4 & 0.007 & 0.109 & 0.249 & $-$0.49 & $-$0.72 & ch1 cl25 & 0.001 & 0.324 & 0.435 & $-$0.32 & $-$0.48 \\
ch0 cl5 & 0.007 & 0.265 & 0.214 & $-$0.49 & $-$0.75 & ch1 cl26 & 0.012 & 0.371 & 0.521 & $-$0.14 & $-$0.23 \\
ch0 cl6 & 0.014 & 0.275 & 0.436 & $-$0.62 & $-$1.04 & ch1 cl27 & 0.002 & 0.335 & 0.580 & +0.47 & +0.70 \\
ch0 cl7 & 0.008 & 0.225 & 0.328 & $-$1.04 & $-$2.69 & ch1 cl28 & 0.003 & 0.084 & 0.202 & $-$0.27 & $-$0.33 \\
ch0 cl8 & 0.003 & 0.042 & 0.155 & $-$0.14 & $-$1.06 & ch1 cl29 & 0.002 & 0.176 & 0.303 & +0.07 & +0.44 \\
ch0 cl9 & 0.005 & 0.259 & 0.547 & $-$0.35 & $-$0.74 & ch1 cl30 & 0.004 & 0.075 & 0.123 & $-$0.28 & $-$0.04 \\
ch0 cl10 & 0.000 & 0.067 & 0.238 & $-$0.19 & $-$0.40 & ch1 cl31 & 0.002 & 0.142 & 0.430 & $-$0.70 & $-$0.86 \\
ch0 cl11 & 0.000 & 0.069 & 0.121 & $-$0.32 & $-$0.86 & ch1 cl32 & 0.001 & 0.539 & 0.687 & $-$0.05 & +0.26 \\
ch0 cl12 & 0.003 & 0.162 & 0.202 & $-$0.31 & $-$0.58 & ch1 cl33 & 0.001 & 0.322 & 0.461 & +0.08 & +0.57 \\
ch0 cl13 & 0.011 & 0.229 & 0.144 & $-$0.43 & $-$0.71 & ch1 cl34 & 0.022 & 0.595 & 0.677 & $-$0.17 & $-$0.32 \\
ch0 cl14 & 0.005 & 0.417 & 0.092 & $-$0.52 & $-$0.93 & ch1 cl35 & 0.003 & 0.358 & 0.499 & $-$0.17 & $-$0.22 \\
ch1 cl15 & 0.002 & 0.014 & 0.018 & +0.29 & $-$0.15 & ch1 cl36 & 0.002 & 0.236 & 0.408 & $-$0.19 & $-$0.33 \\
ch1 cl16 & 0.034 & 0.415 & 0.649 & +0.31 & +0.47 & ch1 cl37 & 0.008 & 0.331 & 0.536 & $-$0.17 & $-$0.17 \\
ch1 cl17 & 0.005 & 0.138 & 0.264 & +0.42 & +0.42 & ch1 cl38 & 0.005 & 0.480 & 0.541 & +0.21 & +0.60 \\
ch1 cl18 & 0.008 & 0.109 & 0.296 & $-$0.02 & $-$0.07 & ch1 cl39 & 0.005 & 0.185 & 0.237 & $-$0.00 & +0.29 \\
ch1 cl19 & 0.012 & 0.295 & 0.473 & $-$0.10 & $-$0.29 & ch1 cl40 & 0.017 & 0.576 & 0.562 & $-$0.18 & $-$0.06 \\
ch1 cl20 & 0.007 & 0.353 & 0.485 & $-$0.05 & $-$0.24 & ch1 cl41 & 0.000 & 0.093 & 0.104 & $-$0.03 & +0.26 \\
\midrule
mean & 0.007 & 0.245 & 0.361 & 0.28 & 0.53 & & & & & & \\
\bottomrule
\end{tabular*}

\end{table}

\begin{figure}[tp]
\centering
\includegraphics[width=\textwidth]{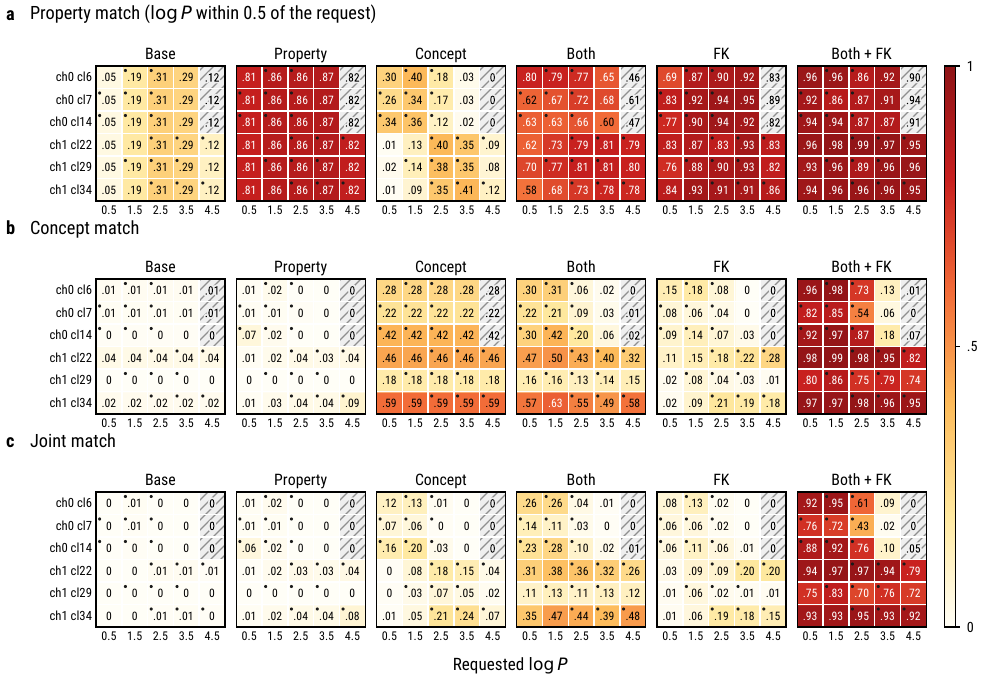}
\caption{\textbf{Joint conditioning on molecular graphs, per target cell.} Share of the generated molecules
that \textbf{a}~have a $\log P$ within $0.5$ of the request, \textbf{b}~are core members of the requested
concept and \textbf{c}~satisfy both, for the six arms of Table~\ref{tab:e4} (columns), per concept (rows) and
requested $\log P$. A dot marks a native cell, and hatching a cell without any training molecule with that
combination. The joint arms hold $1024$ molecules per cell, whereas the base (one sample), the property adapter (one sample per level)
and the concept adapter (one sample per concept) are graded against every cell that they apply to, so that their cells
share molecules. All panels share one color scale.}
\label{fig:g-heat-mol}
\end{figure}

\paragraph{Joint conditioning.}
Table~\ref{tab:g-e4} extends Table~\ref{tab:e4}. The composed adapters with steering reach a higher joint match than every other
arm in all $27$ supported cells. Steering alone, in contrast, needs a median of $128$ samples per distinct joint match, against
$4.2$ for the composed adapters with steering, and it finds none in two of the $27$ supported cells. The best arm differs
strongly between the two channels (Figure~\ref{fig:g-heat-mol}), with a joint match of $0.48$ for the three concepts of
channel 0 and of $0.87$ for the three concepts of channel 1, averaged over their cells. The concepts of channel 0 reach
between $0.76$ and $0.92$ at $\log P = 0.5$ and between $0.02$ and $0.10$ at $3.5$. This agrees with the training data, in
which the cells at the two highest levels are off-native or unsupported for all three of them. However, the spread of the
best arm over the cells remains large ($0.676 \pm 0.344$), and the sampling error of a single cell of an arm without steering is up to $0.03$ at $1024$
molecules.

\begin{table}[tp]
\centering
\footnotesize
\setlength{\tabcolsep}{1.5pt}
\caption{\textbf{Joint conditioning on molecular graphs, with the spread over the cells.}
Table~\ref{tab:e4} with the standard deviation over the $30$ target cells as small numbers, and over the
cells of the respective class for the native and the off-native columns. A dash marks an arm without a
distinct joint match in more than half of the $27$ supported cells, and the last column counts the supported
cells without one.}
\label{tab:g-e4}
\begin{tabular*}{\linewidth}{@{\extracolsep{\fill}} l ccc cc cc r c}
\toprule
 & \multicolumn{3}{c}{Match $\uparrow$} & \multicolumn{2}{c}{Joint by cell $\uparrow$} & & & & \\
\cmidrule(lr){2-4}\cmidrule(lr){5-6}
Model & Prop. & Concept & Joint & Native & Off-nat. & Unique $\uparrow$ & Valid $\uparrow$ & Cost $\downarrow$ & No hit $\downarrow$ \\
\midrule
Base (no conditioning) & 0.192{\scriptsize$\pm$.101} & 0.014{\scriptsize$\pm$.012} & 0.003{\scriptsize$\pm$.004} & 0.004{\scriptsize$\pm$.004} & 0.001{\scriptsize$\pm$.001} & \textbf{1.000}{\scriptsize$\pm$.000} & 0.988{\scriptsize$\pm$.000} & 1024.0 & 8/27 \\
$+$ property adapter & 0.844{\scriptsize$\pm$.025} & 0.017{\scriptsize$\pm$.022} & 0.015{\scriptsize$\pm$.020} & 0.022{\scriptsize$\pm$.023} & 0.007{\scriptsize$\pm$.008} & \textbf{1.000}{\scriptsize$\pm$.000} & 0.946{\scriptsize$\pm$.011} & 113.8 & 3/27 \\
$+$ concept adapter & 0.183{\scriptsize$\pm$.152} & 0.357{\scriptsize$\pm$.148} & 0.067{\scriptsize$\pm$.074} & 0.102{\scriptsize$\pm$.076} & 0.028{\scriptsize$\pm$.041} & \textbf{1.000}{\scriptsize$\pm$.000} & 0.978{\scriptsize$\pm$.008} & 19.0 & 2/27 \\
$+$ both adapters & 0.698{\scriptsize$\pm$.098} & 0.264{\scriptsize$\pm$.199} & 0.194{\scriptsize$\pm$.155} & 0.223{\scriptsize$\pm$.139} & 0.203{\scriptsize$\pm$.171} & \textbf{1.000}{\scriptsize$\pm$.001} & 0.948{\scriptsize$\pm$.029} & 4.4 & \textbf{0}/27 \\
$+$ FK ($k{=}16$) & 0.871{\scriptsize$\pm$.061} & 0.091{\scriptsize$\pm$.080} & 0.064{\scriptsize$\pm$.066} & 0.096{\scriptsize$\pm$.069} & 0.031{\scriptsize$\pm$.032} & 0.338{\scriptsize$\pm$.044} & \textbf{0.999}{\scriptsize$\pm$.001} & 128.0 & 2/27 \\
$+$ both adapters $+$ FK & \textbf{0.932}{\scriptsize$\pm$.038} & \textbf{0.719}{\scriptsize$\pm$.344} & \textbf{0.676}{\scriptsize$\pm$.344} & \textbf{0.813}{\scriptsize$\pm$.145} & \textbf{0.639}{\scriptsize$\pm$.402} & 0.365{\scriptsize$\pm$.069} & 0.992{\scriptsize$\pm$.006} & \textbf{4.2} & \textbf{0}/27 \\
\bottomrule
\end{tabular*}

\end{table}

\subsection{Real-World Mutagenicity}
\label{app:perconcept:realworld}

Table~\ref{tab:g-e5} reports the six arms of Experiment~4 on the $14$ cells of \appref{app:protocol:mutagenicity}, which
request each of seven concepts for both classes, with the columns of Table~\ref{tab:e4}. The concepts are the N-nitrosamines
(ch1 cl8), the epoxides (ch1 cl10), the azides (ch1 cl6), the nitro-heteroarenes (ch1 cl24), the azo compounds (ch1 cl17), the
aromatic amines (ch1 cl20) and the sulfonyl compounds (ch0 cl1). The grader of this rung is the \megan{} model that also supplies
the energy of steering, and we therefore read every rate as the extent to which an arm steers the prediction of this model. Under
this reading, the composed adapters with steering reach a joint match of $0.592$, against $0.220$ for the composed adapters alone
and $0.004$ for the base, and they are ahead of all other arms in each of the nine supported cells. The price is the diversity of
the samples, since the share of distinct graphs falls from $0.638$ for the composed adapters to $0.164$ with steering. An arm
without the class adapter is graded at both classes, so that its joint match is half its concept match. This makes the concept
adapter alone ($0.261$) look better than the composed adapters ($0.220$), and we therefore do not compare these two arms.

\begin{table}[tp]
\centering
\footnotesize
\setlength{\tabcolsep}{1.5pt}
\caption{\textbf{Joint conditioning on mutagenicity, with the spread over the cells.} The six arms on the
$14$ cells of seven concepts and two classes, with the columns of Table~\ref{tab:e4} and the standard
deviation over the cells as small numbers (the off-native column holds a single cell). The class adapter is
used at $w = 1.5$, the concept adapter at $w = 1$ and the composition at $(w_c, w_p) = (1, 1)$. The cost is
the median over the $9$ supported cells, and the last column counts the supported cells without a distinct
joint match.}
\label{tab:g-e5}
\begin{tabular*}{\linewidth}{@{\extracolsep{\fill}} l ccc cc cc r c}
\toprule
 & \multicolumn{3}{c}{Match $\uparrow$} & \multicolumn{2}{c}{Joint by cell $\uparrow$} & & & & \\
\cmidrule(lr){2-4}\cmidrule(lr){5-6}
Model & Prop. & Concept & Joint & Native & Off-nat. & Unique $\uparrow$ & Valid $\uparrow$ & Cost $\downarrow$ & No hit $\downarrow$ \\
\midrule
Base (no conditioning) & 0.495{\scriptsize$\pm$.365} & 0.007{\scriptsize$\pm$.011} & 0.004{\scriptsize$\pm$.008} & 0.006{\scriptsize$\pm$.010} & 0.006 & \textbf{0.999}{\scriptsize$\pm$.000} & 0.990{\scriptsize$\pm$.000} & 512.0 & 4/9 \\
$+$ class adapter & 0.750{\scriptsize$\pm$.088} & 0.008{\scriptsize$\pm$.008} & 0.007{\scriptsize$\pm$.009} & 0.011{\scriptsize$\pm$.009} & 0.003 & 0.920{\scriptsize$\pm$.037} & 0.939{\scriptsize$\pm$.024} & 102.4 & \textbf{0}/9 \\
$+$ concept adapter & 0.467{\scriptsize$\pm$.367} & 0.521{\scriptsize$\pm$.144} & 0.261{\scriptsize$\pm$.267} & 0.434{\scriptsize$\pm$.227} & 0.086 & 0.707{\scriptsize$\pm$.108} & 0.933{\scriptsize$\pm$.059} & \textbf{3.5} & \textbf{0}/9 \\
$+$ both adapters & 0.456{\scriptsize$\pm$.354} & 0.447{\scriptsize$\pm$.181} & 0.220{\scriptsize$\pm$.239} & 0.362{\scriptsize$\pm$.226} & 0.054 & 0.638{\scriptsize$\pm$.104} & 0.883{\scriptsize$\pm$.095} & 5.8 & \textbf{0}/9 \\
$+$ FK ($k{=}16$) & \textbf{0.976}{\scriptsize$\pm$.016} & 0.070{\scriptsize$\pm$.115} & 0.069{\scriptsize$\pm$.115} & 0.108{\scriptsize$\pm$.141} & 0.043 & 0.437{\scriptsize$\pm$.132} & \textbf{0.996}{\scriptsize$\pm$.002} & 146.3 & 2/9 \\
$+$ both adapters $+$ FK & 0.813{\scriptsize$\pm$.273} & \textbf{0.755}{\scriptsize$\pm$.229} & \textbf{0.592}{\scriptsize$\pm$.383} & \textbf{0.834}{\scriptsize$\pm$.256} & \textbf{0.636} & 0.164{\scriptsize$\pm$.051} & 0.991{\scriptsize$\pm$.012} & 11.5 & \textbf{0}/9 \\
\bottomrule
\end{tabular*}

\end{table}

The unsupported cells illustrate this reading (Figure~\ref{fig:g-heat-rw}). At the mutagenic class, steering with both adapters
reaches a joint match of $0.54$ for the sulfonyl compounds (ch0 cl1) and of $0.29$ for the azo compounds (ch1 cl17), although
fewer than ten of their core members have this class in the data. At the non-mutagenic class, the joint match stays at $0.07$ for
the azides (ch1 cl6) and at $0.04$ for the N-nitrosamines (ch1 cl8). We believe that the first two rates measure how far the
sampler can move the prediction of the model and do not show that such molecules exist. In the other cells, the joint match of the best arm
lies between $0.96$ and $0.98$ in the five native cells of the mutagenic class, and between $0.35$ and $0.97$ in the native and
off-native cells of the non-mutagenic class. Each cell holds $1024$ graphs for the joint arms, so that the sampling error of a
single rate of an arm without steering is up to $0.03$.

\begin{figure}[tp]
\centering
\includegraphics{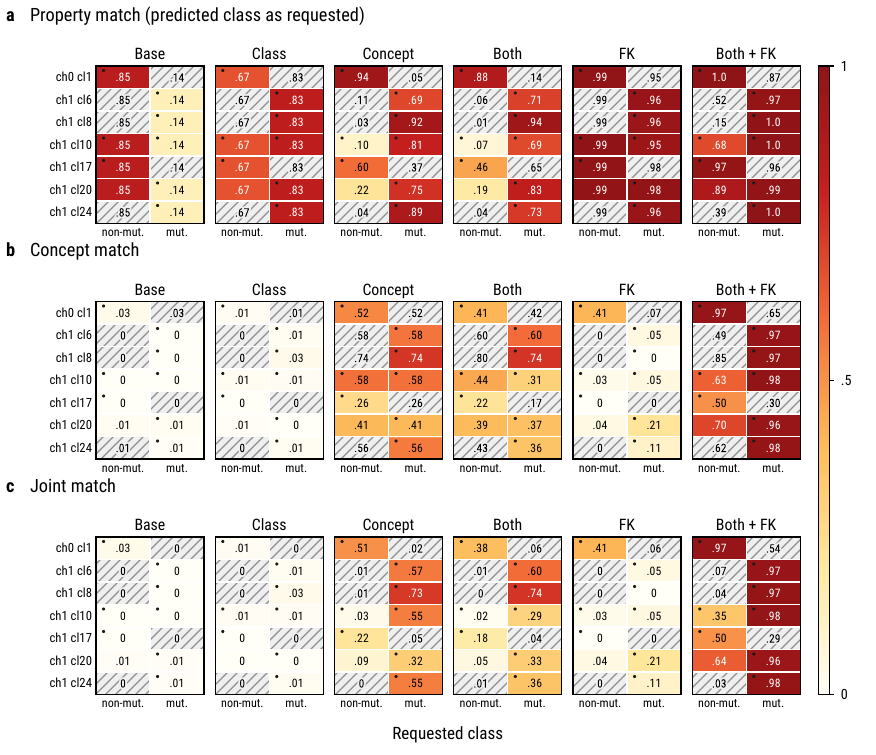}
\caption{\textbf{Joint conditioning on mutagenicity, per target cell.} Share of the generated molecules
that \textbf{a}~have the requested class as predicted by \megan{}, \textbf{b}~are core members of the
requested concept and \textbf{c}~satisfy both, for the six arms of Table~\ref{tab:g-e5} (columns), per concept
(rows) and requested class. A dot marks a native cell, and hatching an unsupported one. The joint arms hold $1024$
molecules per cell, whereas the base ($1024$ in total), the class adapter ($1024$ per class) and the concept adapter ($1024$
per concept) are graded against every cell that they apply to, so that their cells share molecules. The arms
without the class adapter are graded at both classes. All panels share one color scale.}
\label{fig:g-heat-rw}
\end{figure}

\renewcommand{\topfraction}{0.95}\renewcommand{\textfraction}{0.05}\renewcommand{\floatpagefraction}{0.5}
\subsection{Quality of the MEGAN Models}
\label{app:perconcept:megan}

Every concept, every membership value and, on the real-world rung, every property match is read from a \megan{}
model that we train on the data of its rung, so that the quality of this model limits what a rate of the paper
can mean. Table~\ref{tab:megan-quality} lists the three models on test graphs that no step of their training has used.

\begin{table}[!ht]
\centering
\small
\setlength{\tabcolsep}{3.5pt}
\caption{Quality of the \megan{} models on their test graphs. The prediction columns give the coefficient of
determination and the mean absolute error for the two regression models, and the accuracy and the macro F1 score for
the classifier. The explanation AUC and the fidelity sign are given per channel, and the last column is the number of
concepts that the clustering of the explanation embeddings yields.}
\label{tab:megan-quality}
\begin{tabular*}{\textwidth}{l @{\extracolsep{\fill}} r l cc cc r}
\toprule
 & & & \multicolumn{2}{c}{Explanation AUC} & \multicolumn{2}{c}{Fidelity sign} & \\
\cmidrule(lr){4-5}\cmidrule(lr){6-7}
Model & Test graphs & Prediction & Ch.~0 & Ch.~1 & Ch.~0 & Ch.~1 & Concepts \\
\midrule
Colored graphs & $5000$ & $R^2$ $0.954$, MAE $0.333$ & $0.892$ & $0.896$ & $0.73$ & $0.74$ & $19$ \\
Crippen $\log P$ & $1000$ & $R^2$ $0.981$, MAE $0.123$ & $0.953$ & $0.968$ & $0.99$ & $0.96$ & $42$ \\
Ames mutagenicity & $600$ & Acc.\ $0.760$, F1 $0.758$ & $0.807$ & $0.829$ & $0.34$ & $0.82$ & $29$ \\
\bottomrule
\end{tabular*}
\end{table}

The explanation AUC is the AUC of the approximation task of the co-training (\appref{app:megan}), in which the
pooled importance of a channel alone has to tell whether a graph lies on the side of the target that the channel
stands for. The fidelity sign is the share of graphs for which a channel moves the prediction in the direction that
it stands for. A model passes if both lie above $0.6$ and $0.5$ respectively, and both checks warn below $0.7$.
For the two molecular models, we chose the sparsity of the explanations on a validation set and read the test graphs once.

The two regression models predict their targets well and pass every check. The fidelity sign of the colored graphs is
the lower of the two, at $0.73$ and $0.74$, and it still lies above the level at which the check warns. The
mutagenicity model is weaker. Its accuracy lies below the $0.809$ of a random forest on random splits of the same
graphs, and the fidelity sign of its non-mutagenic channel is $0.34$, which fails the check, since this class has no
positive structural signature that leaving out nodes could remove. We therefore read the rates of the real-world rung
as comparisons between arms that one model grades (\Secref{app:protocol:mutagenicity}).

\ifsipart
\bibliographystyle{unsrtnat}
\bibliography{main}
\fi
\fi %

\end{document}